\documentclass[10pt]{article}

\usepackage[
  letterpaper,
  left=1in,
  right=1in,
  top=1in,
  bottom=1in
]{geometry}

\usepackage{setspace}
\usepackage[T1]{fontenc}
\usepackage{lmodern}      
\usepackage{microtype}
\usepackage{tabularx} 
\usepackage{booktabs}  
\usepackage{multirow}  

\usepackage{titlesec}

\titleformat{\section}
  {\normalfont\large\bfseries}{\thesection}{1em}{}
\titleformat{\subsection}
  {\normalfont\normalsize\bfseries}{\thesubsection}{1em}{}
\titleformat{\subsubsection}
  {\normalfont\normalsize\itshape}{\thesubsubsection}{1em}{}

\titlespacing*{\section}{0pt}{2.5ex plus 1ex minus .2ex}{1.5ex}
\titlespacing*{\subsection}{0pt}{2ex plus 0.8ex minus .2ex}{1ex}

\usepackage{graphicx}
\usepackage{booktabs}
\usepackage{multirow}
\usepackage{xcolor}

\usepackage[style=numeric-comp, sorting=none, backend=biber]{biblatex}
\usepackage[acronym]{glossaries}
\makeglossaries

\newcommand{\rev}[1]{\textcolor{black}{#1}}
\newenvironment{revblock}{\color{black}}{}

\begin{document}

\title{Towards Accessible Robot Control: Comparing Kinesthetic Teaching,
SpaceMouse Teleoperation, and a Mixed Reality Interface}

\author{
Aliyah Smith \\
Stanford University \\
\texttt{aliyah1@stanford.edu}
\and
Monroe Kennedy III \\
Stanford University \\
\texttt{monroek@stanford.edu}
}
\date{}

\maketitle
\glsdisablehyper
\newacronym{lfd}{LfD}{learning from demonstration}
\newacronym{vr}{VR}{virtual reality}
\newacronym{ar}{AR}{augmented reality}
\newacronym{mr}{MR}{mixed reality}
\newacronym{dof}{DOF}{degree-of-freedom}
\newacronym{tlx}{TLX}{Task Load Index}
\newacronym{sus}{SUS}{System Usability Scale}
\newacronym{ai}{AI}{artificial intelligence}

\begin{abstract}
  
\rev{Teleoperation interfaces are essential tools for enabling human control of robotic systems. Although a wide range of interfaces has been developed, a persistent gap remains between the level of performance humans can achieve through these interfaces and the capabilities afforded by direct human-guided robot control. This gap is further exacerbated when users are inexperienced or unfamiliar with the robotic platform or control interface. In this work, we aim to better characterize this performance gap for non-expert users by comparing two teleoperation approaches—SpaceMouse teleoperation and a Mixed Reality (MR) interface—against kinesthetic teaching as a non-teleoperation baseline. All three approaches were evaluated in a comprehensive user study involving two robotic platforms and six complex manipulation tasks. Quantitative results show that the SpaceMouse and \gls{mr} interfaces performed comparably, with significant differences in task completion observed for only two tasks, and success rates declining as task complexity increased. Qualitative analysis reflected these trends, highlighting differences in Physical Demand and identifying interface attributes that influence users’ ability to perform, learn, and understand. This study quantifies the limitations of current teleoperation methods and incorporates subjective feedback from 25 participants. The results highlight the critical need to design and rigorously evaluate teleoperation systems for non-expert users, particularly in contexts where autonomous robots are deployed in personal or everyday environments, to ensure usability, efficiency, and accessibility.}

\end{abstract}

\section*{Keywords}
Robotic Manipulation, Teleoperation, Human–Robot Interaction (HRI),
Mixed Reality (MR)

\section{Introduction}

\begin{revblock}
    
Robot teleoperation has long enabled experts to operate robots in remote, hazardous, and otherwise inaccessible environments \cite{vevi_hine_1995,rustremoval_teleop,searchandrescue_settimi,natural_garca_2017}. In such settings, teleoperation interfaces are commonly tailored to the task and place significant demands on operator skill and domain knowledge, yet even highly trained users frequently struggle during remote operation \cite{stillnotsolved_rea}. Collocated teleoperation, in which the operator and robot occupy the same physical workspace, has emerged as an effective method for collecting high-quality data for learning from demonstration. This arrangement enables skilled users to perform complex, dexterous manipulation tasks that serve as training examples, allowing robots to later reproduce these behaviors autonomously. Across both collocated and remote scenarios, researchers and designers have explored a wide range of interface modalities—including joysticks, 3D SpaceMouses \cite{spacemouse_wireless}, \gls{vr} controllers \cite{deepIL, rosen2018testing}, cameras \cite{autonomous_rakita_2018}, teach pendants \cite{augmented_pan_2021}, and data gloves \cite{robotic_binfang_2015}. Despite this diversity, these systems generally require substantial operator skill and domain knowledge, and many have not been evaluated from a user-centered perspective, leaving the cognitive and physical demands largely unknown.

We are now entering an era in which robotic systems are becoming increasingly prevalent in homes, workplaces, and public spaces, where they will be trained to perform complex and meaningful tasks. The training data—whether collected from skilled operators, online videos, or other sources—will provide a strong foundation for these intelligent systems. Nevertheless, a persistent need for human-in-the-loop involvement remains—to correct undesirable behaviors, supply additional demonstrations, or intervene when robots encounter unforeseen situations \cite{hri_sheridan}. While remote teleoperators currently support deployed systems when they fail, future deployments will require mechanisms that enable collocated, non-expert, everyday users to assume this role effectively.

To this end, examining existing teleoperation methods (designed for expert users) through the lens of non-expert operators is critical, especially for high-degree-of-freedom robots and in complex scenarios. The human–robot interaction community has long developed user-centered teleoperation interfaces and conducted studies with non-expert users, often evaluating performance through both quantitative and qualitative metrics \cite{cyber_walker, singh, vibi_quintero_2015, comprehensive_jiang_2024}. However, these studies frequently focus on robots with low degrees of freedom, large industrial manipulators, or humanoid platforms unlikely to be deployed in homes or offices, or they examine relatively simple tasks that do not fully capture the complexity of future scenarios in which non-experts may be asked to demonstrate more challenging skills.

In this work, our primary contribution is an experimental evaluation of two teleoperation approaches contextualized within a broader robot control framework, based on a comprehensive user study involving non-expert users controlling a 7 \gls{dof} manipulator. Through this study, we aim to identify the specific design elements of these methods—two teleoperation approaches and one non-teleoperation approach—that facilitate or hinder user performance during complex manipulation tasks. The methods selected for this study were motivated by their strong precedent in recent work and prevalence in laboratory settings, yet their limited evaluation in human-centered contexts. SpaceMouse teleoperation, for example, has been widely adopted by robotics researchers for many years; however, to the best of our knowledge, systematic human-centered evaluations have not been conducted for the device. Similarly, mixed-reality–based interfaces have been examined in a number of human-centered studies, but these evaluations have typically been limited to simple manipulation tasks or unideal robotic platforms. Finally, kinesthetic teaching was included in this study to provide contextual evidence of what is achievable through direct robot control. Despite its effectiveness, this method has been relatively underexplored in human-centered evaluations. Building on Rea et al.'s work on user-centered teleoperation, which discusses challenges and research directions, we adopt a framework that considers how interfaces help the user learn, help the user do, and help the user understand \cite{stillnotsolved_rea}. By isolating and analyzing these design elements, we aim to provide insights into the design of existing teleoperation interfaces, explore their suitability for everyday users operating high–degree-of-freedom manipulators in unstructured environments, and inform future system designs that maximize user understanding, comfort, and performance while lowering the barriers to everyday teleoperation.
\end{revblock}

\section{Related Work}

\begin{revblock}

Teleoperation interfaces have been explored extensively across a wide range of robotic platforms, including humanoids \cite{teleoperation_darvish_2023, humanplushumanoidshadowingimitation, cheng2024opentelevisionteleoperationimmersiveactive, omnih2o, firstperson_fritsche_2015}, mobile and aerial robots \cite{mobileALOHA, robot_walker_2019, vr_stotko_2019, improving_hedayati_2018, utilization_kot_2014}, and robotic manipulators for applications such as remote operation in hazardous or inaccessible environments \cite{worldtrade_casper, immersive_martins_2009}, industrial automation \cite{industrial_prez_2019, digital_kuts_2019, haptic_ni_2017, interactive_zh_2006, teleoperation_galbis_2020}, surgical robotics \cite{multisensor_qi_2021}, and robot learning. In contrast to these broad domains, our work focuses specifically on collocated teleoperation, an area closely tied to robot learning workflows such as data collection, human intervention, and corrective demonstrations. We underscore this distinction because our goal is to understand how non-expert users can operate robots through existing interfaces in everyday settings—such as homes, workplaces, and public environments—where usability, safety, and accessibility are paramount and visual obstruction of the robot and its workspace is less likely. 

\subsection{Teleoperation Interfaces that Enable Complex Manipulation}
Teleoperation systems designed for data collection must be highly capable and sufficiently robust to support long-duration use, often requiring tens to hundreds of demonstrations for effective policy training. Prior work has explored a variety of interface modalities for this purpose. Zhang et al. introduced a \gls{vr}-based teleoperation interface for collecting human demonstrations to train imitation learning policies for manipulation \cite{deep_zhang_2017}. Fu et al. developed MobileALOHA, a whole-body teleoperation system aimed at generating training data for autonomous control policies \cite{mobileALOHA}. DexPilot leverages a vision-based pipeline, using multiple cameras to track human hand poses for controlling a high-\gls{dof} multifingered robot \cite{dexpilot_handa_2020}. Gello et al. proposed a low-cost approach that relies on inexpensive physical replicas of robotic manipulators to enable intuitive operator control \cite{gello}. DexCap provides a hand motion-capture system designed for supplying corrective demonstrations to refine learned policies \cite{dexcap}. Mandlekar et al. employed a smartphone as a motion controller paired with a laptop for camera feedback \cite{humanintheloop_mandlekar_2020}. More recently, Holo-Dex and OPEN TEACH have utilized \gls{vr} headsets for hand and finger tracking, enabling teleoperation of multifingered robotic hands with integrated visual feedback \cite{holodex, openteach}. These systems have made it possible for humans to demonstrate complex, multi-step tasks—including washing dishes, cooking meals, and folding laundry.

Although these systems have been applied extensively in robot-learning contexts—particularly for robots intended to operate in unstructured human environments—they are rarely evaluated in scenarios where non-expert users must teleoperate robots to perform real-world tasks. For instance, a home robot trained to place food in an oven may rely on a non-expert user to demonstrate how to operate an air fryer, or a household member may wish to teach a robot their preferred method for folding clothing. Yet prior work offers little in-depth analysis of how existing interfaces support performance without domain knowledge, maintain user comfort, manage cognitive load, or meet the practical demands of everyday use. Despite these gaps, such systems continue to be widely adopted within the robotics community. This raises an important question: do the features that make these interfaces appealing to researchers also enable effective use by non-expert operators?

\subsection{Human-Centered Teleoperation Interfaces}
On the other hand, numerous teleoperation interfaces have been designed from a user-centric perspective and evaluated through extensive studies involving non-expert participants. However, the tasks used in these evaluations are often simplified or highly constrained, and therefore fail to capture the complexity and variability of demonstrations that users may need to perform in real-world settings. Moreover, many of the robotic platforms employed in these studies are unlikely to operate in everyday environments such as homes or offices, limiting the relevance of these findings for the contexts in which non-expert users are most likely to interact with robots.

Brizzi et al. conducted a usability study with 22 participants to evaluate an \gls{ar}-based interface designed to provide augmented feedback; however, the system was assessed only on simple pick-and-place tasks \cite{effects_brizzi_2018}. Singh et al. introduced a tablet-based teleoperation interface for manipulator control and compared it with a standard commercial keypad, using tasks inspired by real-world scenarios but ultimately restricted to laboratory proxies; moreover, their manipulator featured only four degrees of freedom \cite{singh}. Naceri et al. developed a \gls{vr} interface using HTC Vive Pro motion controllers, yet their evaluation included only six male participants, limiting generalizability \cite{towards_naceri_2019}. Kent et al. proposed two teleoperation interfaces that incorporated scene and object information, conducting a comparison study using both objective and subjective user-centered metrics \cite{comparison_kent_2017}. However, their approach relied on a 2D interface intended for remote teleoperation, which constrains spatial understanding. 

Broader advances in \gls{ar}/\gls{vr}/\gls{mr} technologies have likewise motivated the development of a wide range of immersive teleoperation systems, demonstrating the potential of these modalities to support more user-centric interaction paradigms \cite{augmented_milgram_1995, survey_billinghurst_2015, applications_milgram_1993, telerobotic_milgram_1995, systematic_wonsick_2020, tangi_feick_2020}. Yet it remains an open question whether such systems truly enable non-expert users to perform more challenging, everyday tasks while maintaining acceptable levels of comfort and cognitive load. 

\subsection{Enabling Non-Expert Control of High-Complexity Robots}
How do we bridge the gap between robots that require expert teleoperation during development and the long-term goal of systems that everyday users—without any robotics expertise—can confidently control? Rea et al. identified several current and emerging research directions that begin to address this challenge, a subset of which are outlined below alongside relevant prior work \cite{stillnotsolved_rea}.

\textbf{Predictability:} Dragan et al. demonstrated through mathematical modeling that legibility and predictability, while both essential for seamless human–robot interaction, often exist in tension with one another \cite{legibility_dragan_2013}. Bejczy et al. investigated predictability using “phantom” or virtual robots that executed movements prior to the physical robot, introducing a brief time delay. Their results showed that such predictive motion improved user performance in simple manipulation tasks. They further found that specific visual aids—such as dual-view or stereoscopic displays—were necessary to support effective perception and control in three-dimensional tasks \cite{phantom_bejczy_1990}.

\textbf{Situational and Spatial Awareness:} Chen et al. identified several sources of human performance challenges in teleoperation, including limited image bandwidth, time delays, low frame rates, lack of proprioceptive feedback, mismatched frames of reference, reliance on two-dimensional views, frequent attention switching, and motion-induced effects \cite{human_chen_2007}. They additionally outlined tools and modalities for mitigating these issues, such as gesture-based inputs \cite{leveraging_aleotti_2004}, voice commands \cite{robot_alexandrova_2014, interactive_miner_1994}, and \gls{ar}/\gls{vr}/\gls{mr} displays. Nelsen et al. further explored the use of ecological interfaces that integrate robot pose, environmental maps, and video feeds within a \gls{mr} display, demonstrating improvements in robot controllability and related performance measures \cite{ecological_nielsen_2007}.

\textbf{Visualization and Immersion:} As early as 1986, Myers highlighted how two-dimensional visualizations can fundamentally transform programming practices, demonstrating that programming-by-example enables non-expert users to learn more rapidly \cite{visual_myers_1986}. More recently, visualization has played an increasingly prominent role in robot control interfaces. Makris et al. introduced an \gls{ar}-based system that visualizes robot motion, alerts, and production information; although the interface did not directly support robot control, its visual overlays enhanced users’ sense of immersion \cite{augmented_makris_2016}. Lipton et al. developed a \gls{vr} interface inspired by the homunculus model of the mind, embedding users within a virtual control room that presents sensor feedback and maintains a dynamic mapping between the robot and the operator \cite{baxters_lipton_2018}. Additional work has focused on visualizing safety zones \cite{arbased_hietanen_2020}, camera feeds and viewpoints \cite{visual_kim_1987, improving_hedayati_2018}, and robot paths and trajectories \cite{Li, robot_chong_2009, quintero2018robot}, all of which aim to support improved situational awareness and more intuitive teleoperation.

While numerous other factors—such as shared autonomy \cite{formalizing_dragan_2012}—can enhance the teleoperation experience, we focus on the aspects discussed above, as they are most relevant for non-expert users collocated with a robot in scenarios where the user must maintain full control in real-time.

\subsection{Evaluating Teleoperation Systems for Non-expert Users}
Teleoperation systems for data collection most commonly report quantitative metrics, such as task completion time and success rate. Additional measures include robot idle time \cite{arbased_hietanen_2020}, and failure modes \cite{gello}. While useful, these metrics alone are insufficient for evaluating systems intended for non-expert users. User-centered teleoperation studies often incorporate subjective feedback via Likert scales or open-ended questions, addressing themes such as ergonomics, safety, perceived competence, autonomy, and fluency \cite{arbased_hietanen_2020, robot_walker_2019, autonomous_rakita_2018}. Several studies also employ standardized questionnaires to assess usability and workload, including the \gls{sus} \cite{sus, improving_hedayati_2018, whitney2019comparing} and NASA \gls{tlx} \cite{nasa, whitney2019comparing}. We leverage both quantitative and subjective metrics to evaluate and compare three accessible approaches to robot control, two teleoperation systems and direct manipulation, for non-expert users in complex tasks with 7 \gls{dof} manipulators.

\end{revblock}

\section{Control Interfaces}

\begin{figure}
    \centering
    \includegraphics[width=1.0\linewidth]{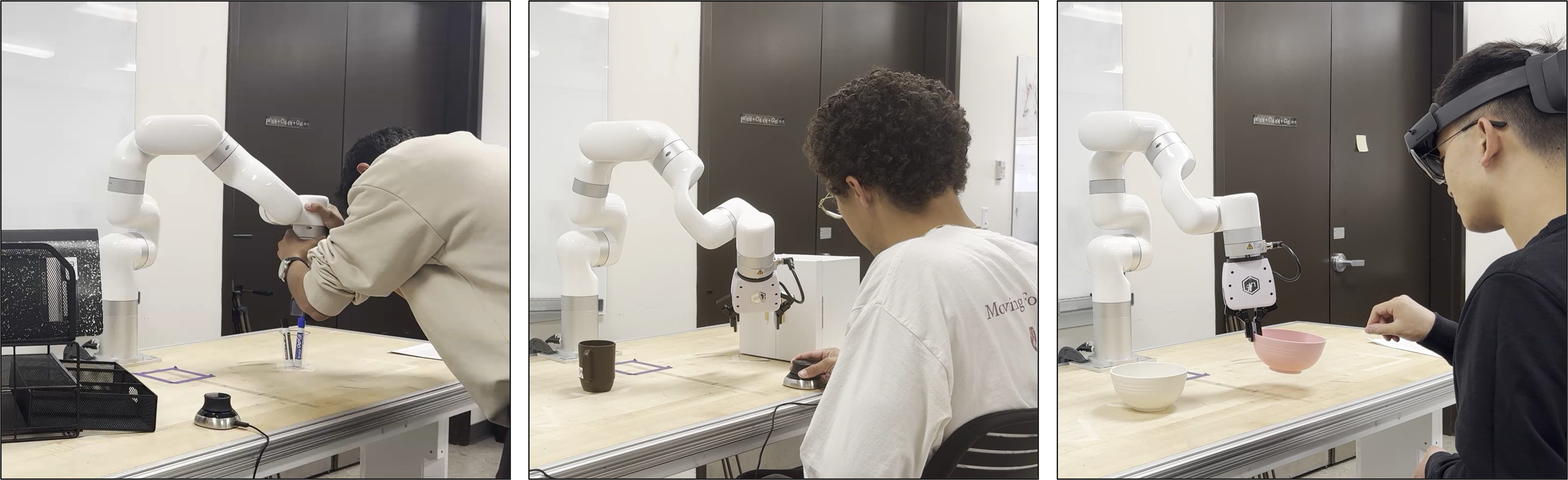}
    \caption{The three control methods investigated in this study. (Left: Kinesthetic Teaching, Center: SpaceMouse Teleoperation, Right: \gls{mr} Teleoperation Interface.)}
    \label{fig:controlmethods}
\end{figure}

\begin{revblock}
In this context, we define accessible control methods as those that are easily transferable across environments, require little or no additional hardware, are intuitive to use, and impose minimal task load. This definition is both necessary and sufficient, as users must be able to control robots effortlessly across multiple locations—for example, moving between a bedroom and a living room within a home. Interfaces that rely on duplicate or environment-specific hardware introduce additional burden, as equipment must be relocated or replicated when changing settings. Usability is a well-established determinant of the adoption of novel technologies; interfaces that are difficult to use create barriers for non-expert users and hinder voluntary adoption. Similarly, task load is a critical factor in sustained use: interfaces that impose high cognitive or physical demands are less likely to be adopted over time or may only be used when absolutely necessary. Collectively, these criteria characterize interfaces that support robust, accessible robot operation across a wide range of everyday contexts.

We focus on general scenarios in which a non-expert user must control a robot in real time to complete complex manipulation tasks; consequently, the methods are not necessarily tailored to a specific use case, such as robot learning or remote teleoperation in hazardous environments. In this study, we compare three approaches—SpaceMouse Teleoperation, a Mixed Reality Teleoperation Interface, and Kinesthetic Teaching—because they satisfy the first two accessibility criteria. We limit the comparison to these three methods due to the constraints of conducting a user study, leaving the evaluation of other approaches, such as motion retargeting, duplicate systems, or alternative \gls{vr}/\gls{ar}/\gls{mr} designs, for future work. Moreover, each system was implemented in its minimal form to better isolate and understand the inherent advantages and limitations of each method; that is, we avoided augmenting systems with features not intrinsic to the approach (e.g., mixed reality headsets inherently support voice commands, whereas the SpaceMouse does not). The following sections provide further details on each method visualized in Figure \ref{fig:controlmethods}, all of which were tested on two robotic platforms: the Kinova Gen3 7 \gls{dof} manipulator and the UFactory xARM 7 \gls{dof} manipulator.
\end{revblock}

\subsection{SpaceMouse Teleoperation} Originally developed by 3Dconnexion for everyday computing and computer-aided design (CAD), the SpaceMouse has proven to be a versatile input device. For years, robotics researchers have adopted it for robot teleoperation due to its effectiveness, compact form factor, portability, and intuitive design. In this study, we use the SpaceMouse Compact to map device inputs to Cartesian velocity commands for both robotic platforms. \rev{The device’s six-degree-of-freedom (\gls{dof}) joystick is mapped directly to end-effector velocities: continuously deflecting the joystick in a given direction produces motion of the end effector in the corresponding direction, as defined by the selected reference frame. Commanded velocities scale proportionally with joystick deflection and return to zero when the joystick is released to its neutral position.}

\begin{figure*}[h]
\centering
\begin{tabular}{lll}
\includegraphics[scale=0.35]{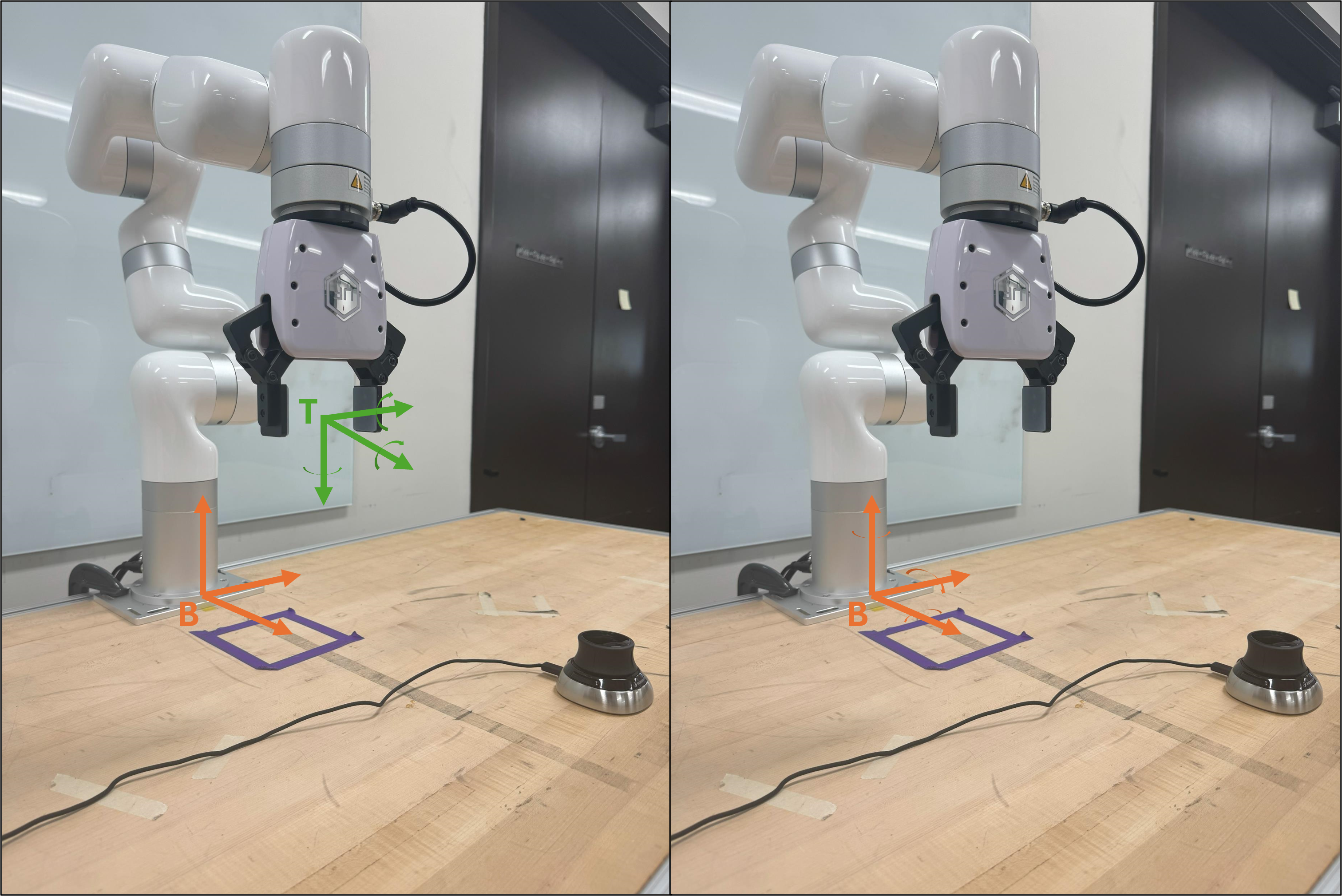}
\end{tabular}
\caption{The mixed reference frame paradigm (left), where SpaceMouse translation inputs are defined in the base (B) frame and rotation inputs in the tool (T) frame, and the base reference frame paradigm (right), where both translation and rotation inputs are defined relative to the base (B) frame. We used the mixed reference frame for the Kinova robot and the base reference frame for the xArm robot.}
\label{fig:spacemouse_frames}
\end{figure*}

For the two robotic platforms, we employ different frames of reference to fully leverage the capabilities of the SpaceMouse, shown in Figure \ref{fig:spacemouse_frames}. For the Kinova Gen3, we adopt the device code from RoboMimic \cite{robomimic2021}, in which velocity commands are defined in a mixed reference frame: translational components are expressed relative to the robot’s base frame, while rotational components are defined in the tool frame. For the UFactory xARM 7, we use the manufacturer’s ROS package for SpaceMouse servoing, which defines all commands with respect to the robot’s base frame. \rev{Readers interested in further details on the control implementation are encouraged to consult the source code.} One advantage of the SpaceMouse is that it supports three frame definitions—base, mixed, and tool—allowing flexibility depending on the context (e.g., third-person versus first-person views). \rev{To avoid constraining our analysis, we test the two most commonly used mappings, using one mapping for one set of tasks, and another mapping for the second set. It is important to note that the overarching control paradigms are consistent across both robotic platforms: inputs consist of velocity commands, and each platform’s inverse kinematics solver converts these Cartesian velocity commands into joint velocities.}

\subsection{Mixed Reality Teleoperation Interface}
\begin{revblock}
Mixed reality enables the overlay of virtual objects onto the real environment \cite{augmented_milgram_1995}. In designing the mixed reality interface for this study, we drew inspiration from a range of prior works to implement a virtual kinesthetic approach employing the following design aspects: digital twins, voice commands, intermittent control \cite{virtual_walker_2023, mixedrealitybased_dasun_2020, assisting_arboleda_2021, connecting_szafir_2021}. See Section 2.3 for additional works that informed and inspired this interface design. While the combination of these design elements may be novel, we do not claim novelty for the design elements themselves. Guided by the principles of kinesthetic teaching, the interface was designed to replicate its intuitive, hands-on style while substantially reducing the physical effort required. Consequently, the \gls{mr} interface prioritizes simplicity and ease of use in its design.
\end{revblock}

\begin{figure*}[h]
\centering
\begin{tabular}{lll}
\includegraphics[scale=0.35]{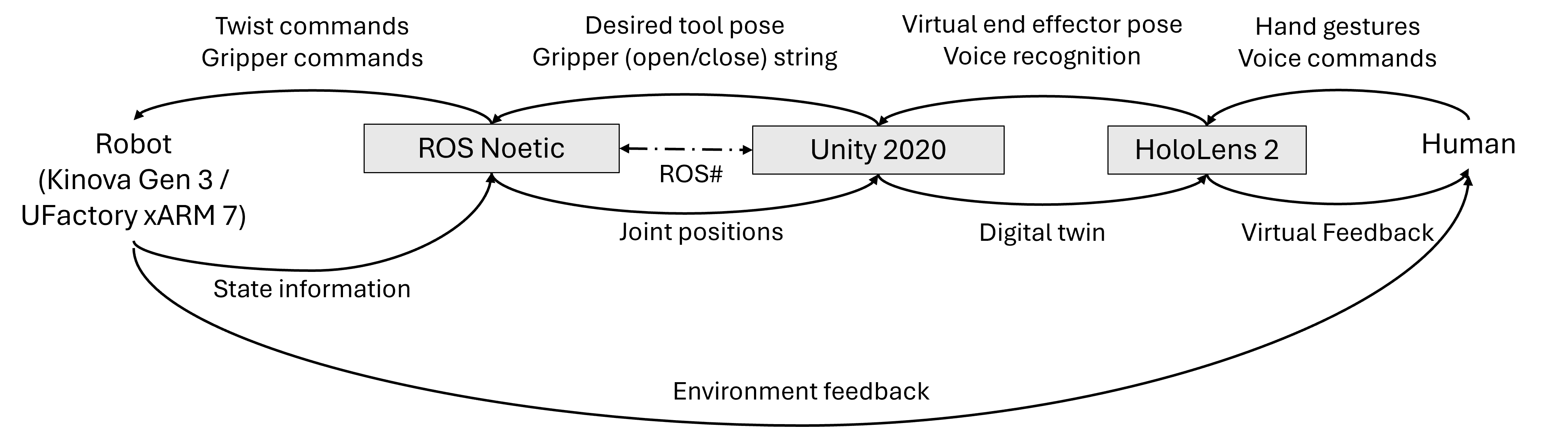}
\end{tabular}
\caption{The \gls{mr} system architecture.}
\label{fig:architecture}
\end{figure*}

\begin{revblock}
\begin{figure}
    \centering
    \includegraphics[width=0.35\linewidth]{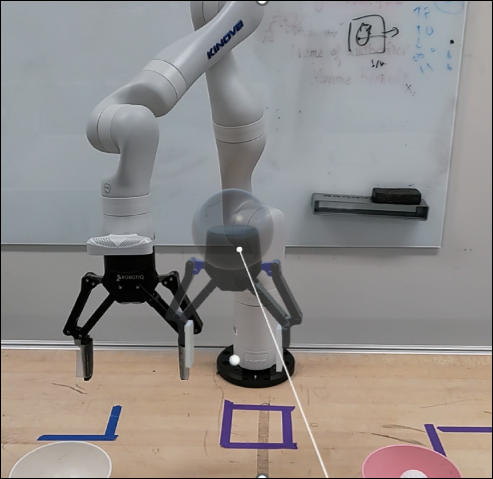}
    \caption{Participants use far interactions (indicated by the white ray coming from the person's hand) to manipulate a virtual sphere and gripper. The physical robot moves in real time to match the end-effector pose defined by the virtual objects.}
    \label{fig:ar_manipulation}
\end{figure}

The overall system architecture of this interface is shown in Figure \ref{fig:architecture}. The system uses the Microsoft HoloLens 2 mixed reality headset, with the application developed in Unity 2020.3 and ROS\# for communication with a Linux system running ROS. After an initial calibration step, users control the robot by manipulating a virtual robot’s end effector, while the physical robot dynamically follows the virtual pose. Users interact via far interactions, pointing a hand ray at the sphere and performing a pinch gesture to engage control. Active engagement in shown through a solid white ray from the user's hand to the virtual object, pictured in Figure \ref{fig:ar_manipulation}. Moving and rotating the hand while pinched manipulates the virtual sphere, which governs the real-time position and orientation of the robot’s end effector. This allows large robot movements with minimal hand motion. The HoloLens tracks the virtual tool frame and sends its pose to ROS, along with a boolean indicating user engagement. During manipulation, the system computes the difference between the robot’s current and target tool poses, capturing linear displacement and rotational offsets (roll, pitch, yaw). These differences are converted into scaled linear and angular velocity commands, which are sent to the robot. The robot’s inverse kinematics solver translates these commands into joint motions, enabling precise and smooth end-effector control. 

The interface supports intermittent control: releasing the virtual sphere stops velocity commands, causing the robot to hold its current pose. Users can fine-tune movements step-by-step. Additional features include a slow mode, which reduces end-effector speed to roughly one-third for more precise manipulation, and multiple gripper control options: voice commands, virtual buttons, or a double-tap hand gesture. 
\end{revblock}
\subsection{Baseline Comparison: Kinesthetic Teaching}
\rev{Following insights from \cite{whitney2019comparing, robot_alexandrova_2014, chen2025dexforceextractingforceinformedactions}, we include kinesthetic teaching, or direct manipulation, as one of the methods for robot control in this study, based on the prediction that it represents best-case performance for manipulation tasks.} Kinesthetic teaching is the most intuitive approach for controlling robotic manipulators, involving physically guiding the robot from an initial configuration to a goal configuration. This control method is typically built into robotic manipulators, with manufacturers providing specific modes to enable kinesthetic teaching (e.g., "Admittance Mode" for the Kinova Gen3 and "Manual Mode" for the UFactory xARM 7), accessible through web interfaces or even physical buttons on the manipulator itself. \rev{We consider this method as the benchmark for objective robot performance against which the teleoperation approaches can be compared. However, comprehensive studies validating this assumption—particularly in comparison with other control methods—are lacking. In other words, we aim to confirm that kinesthetic teaching produces optimal performance and to identify the aspects of this method that enable users to perform at their best, even if doing so may compromise comfort.}

In this work, we use the manufacturers’ manual modes to control the two robotic platforms described above. Because kinesthetic teaching does not provide direct control over the gripper’s open and close actions, alternative methods are employed. For the Kinova Gen3, an Xbox controller is used to operate the gripper, while for the UFactory xARM 7, gripper control is achieved through two buttons on the SpaceMouse.

\rev{Please see the project website for videos of the three control methods \cite{smith2025human_centered_teleop}.}

\begin{revblock}
\subsection{Linking the Control Interfaces to User Learning, Performance, and Understanding}

We selected these three approaches based on their relevance to the key dimensions of user-centered teleoperation: helping users learn, helping users do, and helping users understand \cite{stillnotsolved_rea}.

\textbf{Helping Users Learn:} We hypothesize that the kinesthetic teaching approach enables users to implicitly learn the robot’s underlying kinematics—including joint motions and limitations, its reachable workspace, and potential self-collisions—without requiring explicit instruction. This rapid learning process may contribute to the perception that kinesthetic teaching allows users to achieve optimal performance. However, such performance may come at the cost of substantial physical effort, potentially making it more difficult for users to execute tasks comfortably (Not Helping Users Do).

\textbf{Helping Users Do:} We hypothesize that the SpaceMouse’s minimal design enables users to control the robot’s end-effector seamlessly, allowing them to execute large or precise movements with minimal effort. However, users may find the specific mappings unintuitive, making it difficult for users to match inputs with robot movement (Not Helping Users Learn or Understand).

\textbf{Helping Users Understand:} We hypothesize that visualizing the end-effector through the mixed reality digital twin helps users anticipate the robot’s movements in response to their control inputs. This visualization clarifies the relationship between input and robot motion, potentially enhancing task performance. However, depending on users’ prior experience with \gls{ar}/\gls{vr}/\gls{mr}, the interface may be challenging to use, potentially limiting its effectiveness in helping users learn or perform tasks.

While other platforms also support users in these three categories, we limit our comparison to these three diverse methodologies. We believe they encompass a broad range of teleoperation interfaces, including joystick operation, visual feedback, and manipulation of virtual entities. However, we acknowledge that alternative designs of these interface types may produce different outcomes; evaluation of additional platforms and interfaces is left for future work. The following section details the experimental design and evaluation criteria employed to measure the performance and user experience of the three systems.
\end{revblock}

\section{User Study Design and Protocol}

\begin{revblock}
We recruited participants for a within-subjects experiment to control a 7 \gls{dof} manipulator across a series of challenging manipulation tasks, conducted under IRB protocol 65022. Each session lasted approximately two hours, and participants received a \$25 gift card as compensation.

\subsection{Experimental Procedure}
Upon arrival, participants first completed an informed consent form. Participants then completed a two-question pre-experiment survey collecting demographic information (age range) and prior experience in the following areas: robotic manipulators, \gls{ar}/\gls{vr}/\gls{mr}, SpaceMice, and video games. For each category, participants self-reported their experience by selecting one of four options: Never Used, Occasionally Used, Frequently Used, or Expert, with “Expert” defined as active researchers or individuals who regularly work with that technology. After the survey, the researcher provided a brief overview of the three control methodologies and the assigned tasks (see Section 4.3), along with an introduction to the robotic platform. This introduction included general safety guidelines, such as avoiding self-collisions and collisions with the environment, and maintaining awareness of their hands when physically interacting with the robot. 

Following the introduction, participants began with Method 1. The researcher first provided a brief demonstration, after which participants received five minutes of training during which they practiced a representative task. We determined that five minutes of training per method was sufficient based on the nature of the control interfaces, the initial demonstration provided by the researcher, the researcher’s prior experience with the systems, and prior work \cite{gello}. Many participants indicated that they did not require the full training period, in which case the study proceeded directly to the experimental tasks. Upon completing Method 1, participants filled out two surveys. This procedure was repeated for Method 2, followed by a five-minute break. Participants then completed the same process for Method 3 and concluded the study by completing a post-experiment questionnaire to provide additional feedback.

\textbf{Ordering:} Since many participants had little to no prior experience interacting with robotic systems, we elected to always present Kinesthetic Teaching as the first method. This allowed participants to familiarize themselves both with the robot and with the task objectives, which some participants only fully understood after their initial attempts. While this choice introduces an ordering bias—potentially resulting in slightly lower performance for Kinesthetic Teaching due to initial confusion, and possible fatigue effects for subsequent methods—we determined that this decision was necessary and justified for two reasons. First, Kinesthetic Teaching is not a teleoperation method and therefore does not represent a one-to-one comparison with the other approaches. Instead, it serves as a performance baseline, providing an estimate of the upper bound of achievable performance for each task. Although teleoperation methods are unlikely to reach this level, establishing this benchmark is essential for understanding the magnitude of the performance gap. Second, we expect that any negative bias introduced by initial confusion and any positive bias introduced by reduced fatigue effects will have negligible overall impact. Because the primary goal of this study is to assess the accessibility of teleoperation methods for non-expert users, we prioritized ensuring clear task understanding over fully randomizing method order. Randomization could have confounded performance results with misunderstandings of task objectives.

Methods 2 and 3 were counterbalanced across participants to mitigate ordering effects between the teleoperation conditions. All tasks were performed in a fixed order of increasing difficulty, allowing participants to progressively learn throughout the study. Maintaining a consistent task order across participants was also necessary to support subsequent statistical analysis.

\begin{figure}
    \centering
    \includegraphics[width=1.0\linewidth]{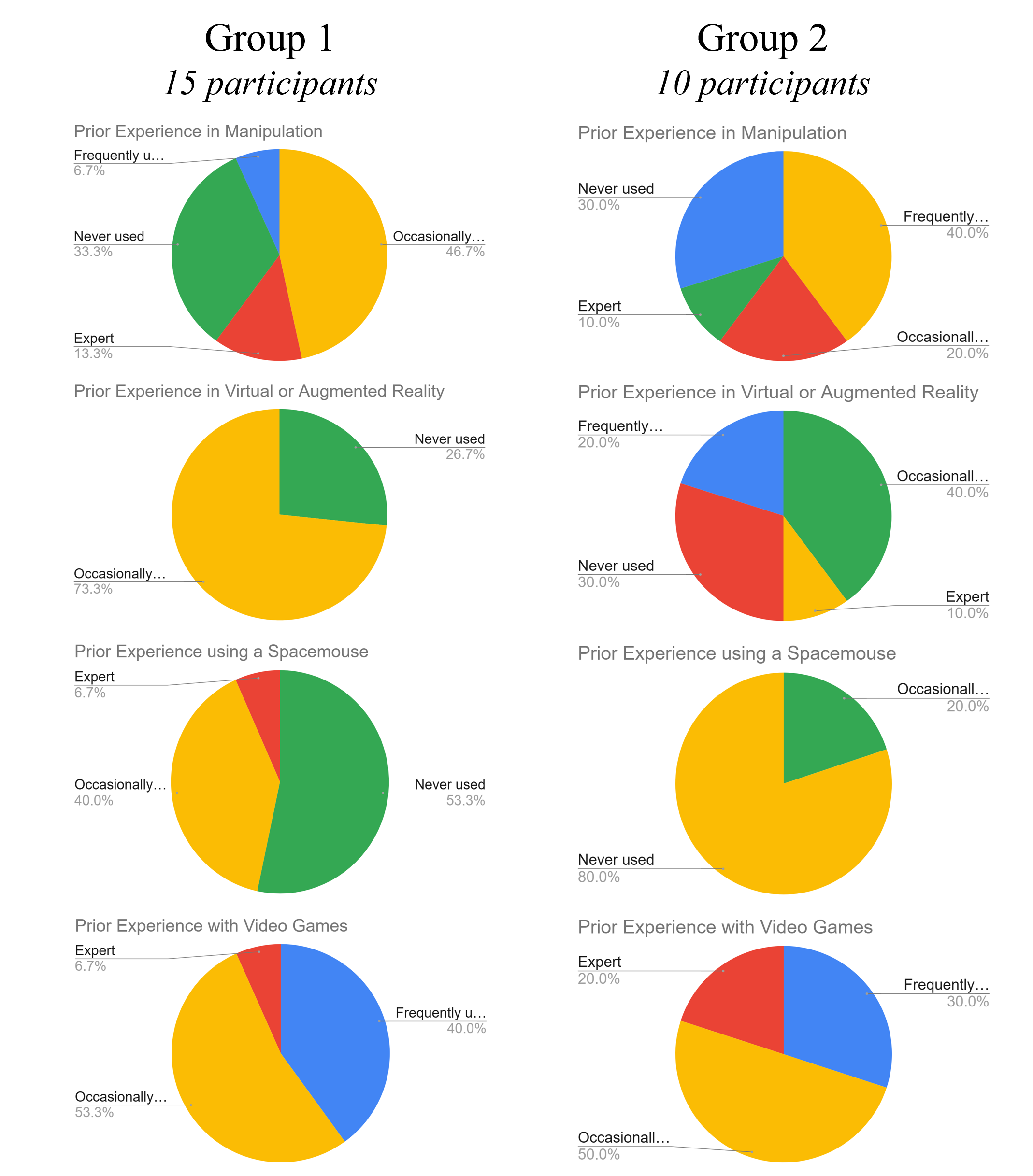}
    \caption{Prior experience across four categories for participants in Groups 1 and 2. Group 1 completed four short-horizon tasks with the Kinova Gen3, while Group 2 completed two long-horizon tasks with the UFactory xARM 7.}
    \label{fig:prior_experience}
\end{figure}

\subsection{Participants}
A total of 25 participants took part in the study. Prior to the experiment, participants were assigned to one of two groups. Group 1 controlled the Kinova Gen3 manipulator on one set of tasks, while Group 2 controlled the UFactory xARM 7 \gls{dof} manipulator on a different set of tasks. We divided participants into two groups to evaluate performance across a broader range of tasks without introducing additional fatigue. 

Accordingly, objective analyses for the two groups are conducted separately, and group demographics are reported independently. The first 15 participants were assigned to Group 1, while the remaining 10 participants were assigned to Group 2 due to timing and resource availability.

Participants in  Group 1 (8 male, 7 female) were in the age range between 18 and 34 (80\% of participants were below the age of 25). Participants in Group 2 (7 male, 3 female) were in the age rage between 18 and 34 (50\% of participants were below the age of 25). Figure \ref{fig:prior_experience} summarizes participants’ prior experience across the surveyed categories for the two groups. Participants who self-reported as having “Never Used” or “Occasionally Used” robotic manipulators are referred to as “non-experts,” as described in the introduction, and make up the majority of participants in this study. We use this term to categorize these participants for the remainder of our analysis. Participants who self-reported as having “Frequently Used” were typically students in robotics-related fields and were considered to possess sufficient domain knowledge to understand the operation and control of robotic manipulators. In Group 1, 80\% of participants were non-experts, while in Group 2, 50\% were non-experts, ensuring a representative sample of non-experts, while also enabling comparisons between participants with and without substantial prior experience.
\end{revblock}

\begin{figure}
    \centering
    \includegraphics[width=1.0\linewidth]{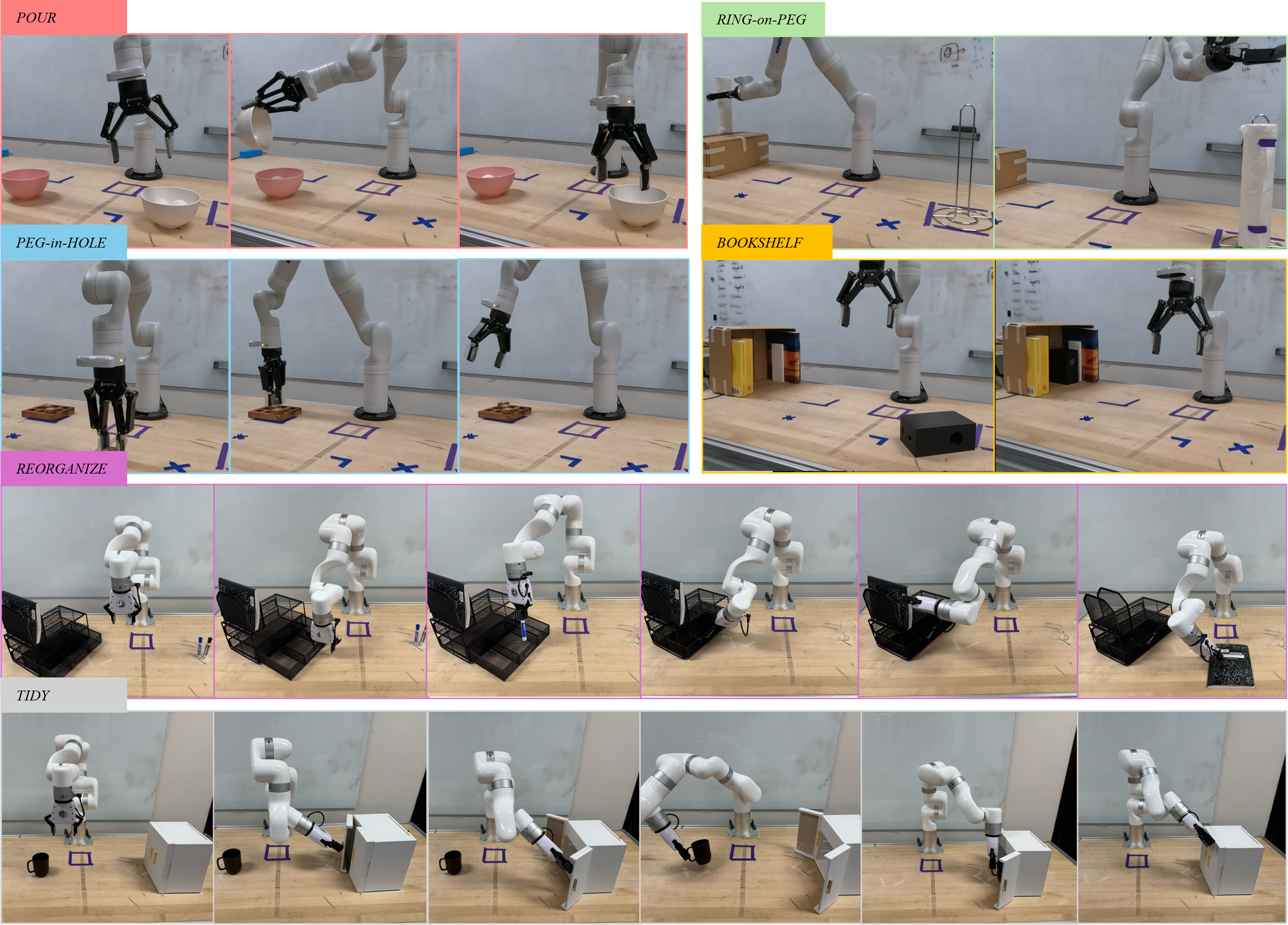}
    \caption{The six tasks included in this study. Group 1 (N=15) completed the short horizon tasks: \textit{POUR}, \textit{PEG-in-HOLE}, \textit{RING-on-PEG}, and \textit{BOOKSHELF}. Group 2 (N=10) completed the long-horizon tasks: \textit{REORGANIZE} and \textit{TIDY.}}
    \label{fig:tasks}
\end{figure}

\begin{table}[ht]
    \centering
    \caption{Breakdown of Participants and Corresponding Tasks}
    \begin{tabularx}{\linewidth}{c c c X}
        \hline
        \textbf{Group} & \textbf{Num. of Participants} & \textbf{Robotic Platform} & \textbf{Tasks Completed} \\
        \hline
        1 & 15 & Kinova Gen3 & \textit{POUR}, \textit{PEG-in-HOLE}, \textit{RING-on-PEG}, \textit{BOOKSHELF} \\
        2 & 10 & UFactory xARM7 & \textit{REORGANIZE}, \textit{TIDY} \\
        \hline
    \end{tabularx}
    \label{tab:group_breakdown}
\end{table}

\subsection{Tasks} We developed six tasks inspired by \cite{comparison_kent_2017} and \cite{wu2024tidybot}, designed to reflect challenges commonly encountered in real-world environments. Four tasks were short-horizon, while two were more complex, multi-step tasks. This design allowed us to assess participant performance on brief, skill-specific tasks as well as their ability to recover and succeed in longer-horizon tasks involving constrained or articulated objects. Group 1 completed the four short-horizon tasks with the Kinova robot, while Group 2 completed the two long-horizon tasks with the xArm, summarized in Table \ref{tab:group_breakdown}. \rev{We chose to use two different robotic platforms to account for potential platform-specific effects that could influence subjective outcomes. For example, moving the Kinova Gen3 requires more force than the xArm, and the kinematics of the two platforms differ, which may cause participants to encounter singularities more frequently on one platform than the other. Such experiences could, in turn, affect perceived usability and task workload.}

\textbf{Short Horizon Tasks:} The first group of participants completed four tabletop manipulation tasks: \textit{POUR}, \textit{PEG-in-HOLE}, \textit{RING-on-PEG}, and \textit{BOOKSHELF}. These tasks were chosen to represent well-known challenges in robotic manipulation while also mimicking practical household activities. Each task was specifically designed to develop and assess critical skills that are difficult yet fundamental to a wide range of manipulation scenarios beyond this study’s immediate focus. Detailed descriptions of the tasks follow, and Figure \ref{fig:tasks} provides a visual overview. At the beginning of each trial, the Kinova arm was placed in a predefined ``top-down” grasp pose to ensure consistency. Each trial was limited to a maximum duration of three minutes.
\begin{itemize}
    \item \textit{POUR}: This task involves two bowls placed on the table—one filled with three ping-pong balls and the other empty. The goal is to lift the bowl containing the balls and carefully pour them into the empty bowl, simulating the act of pouring cereal into a bowl. The task is considered complete when all three balls have been successfully transferred and both bowls are placed back on the table. This task demands smooth, controlled movements to avoid spilling, emphasizing precision and gentle handling.
    \item \textit{PEG-in-HOLE}: The user is required to pick up a tic-tac-toe piece and accurately place it into a specified slot on the tic-tac-toe board. This task highlights fine and precise manipulation skills, as the piece must be aligned correctly within the grid, similar to playing an actual tic-tac-toe game. It exemplifies the classical peg-in-hole manipulation problems often addressed in robotics.
    \item \textit{RING-on-PEG}: Starting with a paper towel roll positioned on an elevated platform on the table and its holder placed opposite it, the objective is to pick up the roll and carefully place it onto the holder. Like the PEG-in-HOLE task, this requires precise alignment and careful positioning, emphasizing dexterity and accurate tool handling to avoid misplacement.
    \item \textit{BOOKSHELF}: This task features a makeshift bookshelf made from a cardboard box set on its side, containing several books. A rigid, 3D-printed object representing a book is placed on the table. The goal is to pick up the object and place it inside the bookshelf, positioning it neatly between the existing books with the circular side facing outward. This task focuses on precision, requiring orientation adjustments and regrasping to correctly place the object within the confined space of the shelf.
\end{itemize}

\textbf{Long Horizon Tasks:} The second group of participants completed two complex, multi-step tabletop manipulation tasks: \textit{REORGANIZE} and \textit{TIDY}. These tasks were designed to simulate more demanding real-world scenarios, such as putting away dishes or organizing an office workspace, which often involve handling articulated, delicate, or soft objects. Each long-horizon task consists of multiple shorter sub-tasks, detailed below. At the start of each trial, the xARM was positioned in its designated ``home" configuration. Each trial was limited to a maximum duration of seven minutes.

\begin{itemize}
    \item \textit{REORGANIZE}: This task begins with a desk organizer placed on the table. The drawer is opened slightly for easy access and a notebook is placed in the folder slot. Two whiteboard markers are placed in a plastic cup on the opposite side of the table. The goal is to complete the following series of sub-tasks: open the drawer, pick \& place the markers in the drawer, close the drawer, and pick \& place the folder onto the table in front of the robot with the stem of the notebook closest to the organizer. This task incorporates the following challenges: opening and closing the drawer (articulated object), grasping the markers (small clearance and offset orientation of the objects), and moving the notebook (constrained space). 
    \item \textit{TIDY}: In this task, a mug and a miniature cabinet are placed on the table. The objective is to open both doors of the cabinet, pick up \& place the mug inside of the cabinet, and then close the doors. Again, working with the articulated object and moving within a confined space make this particular task challenging. 
\end{itemize}

\rev{In this study, the key distinction between short-horizon and long-horizon tasks lies in their evaluation: long-horizon tasks are assessed based on overall task completion, which increases the difficulty of achieving complete success and correspondingly raises the likelihood of partial success. Distinguishing between a series of short-horizon tasks and a single long-horizon task enables analysis of participants’ ability to recover from errors and make strategic choices to achieve the overarching goal. This framing also reinforces the importance of completing all subtasks for participants.}

\subsection{Objective Measures}
During each trial, we recorded task completion time and success. \rev{Other objective metrics, including hand-tracking failures and calibration overhead, were intentionally excluded, as they were not applicable across all three methods.} For time analysis, we report the duration of the most successful trial for each task; if both trials were unsuccessful, the maximum allotted time was used (180 s for short-horizon tasks and 420 s for long-horizon tasks). \rev{For each task, we counted the number of complete successes, partial successes, and failures (2 max for each participant). We report these results aggregated across all participants for each task.} One participant in Group 1 was unable to attempt the final task using the SpaceMouse; therefore, their Bookshelf task data were excluded across all three interfaces.

\subsection{Subjective Measures}
The subjective measures in the study include the NASA \gls{tlx} and the \gls{sus}, \rev{which were administered after participants completed all the tasks for a given interface (3 times in total)}. The NASA \gls{tlx} is a well-established tool widely used to assess workload across various domains \cite{nasa}. It consists of six sub-scales—Mental Demand, Physical Demand, Temporal Demand, Performance, Effort, and Frustration—each rated on a scale from 0 to 100 in 5-point increments. The overall task load is computed by averaging the scores of these sub-scales. Additionally, the \gls{sus} is a commonly used ten-item Likert scale questionnaire that evaluates system usability, producing scores ranging from 0 to 100 \cite{sus}. \rev{Additionally, a post-experiment survey, developed by the authors, consisted of open-ended questions to capture further feedback on each approach and participants’ overall preferences.}

\subsection{Statistical Analyses}
\rev{Data were analyzed using a repeated measures ANOVA, followed by post-hoc paired t-tests for all measures, regardless of normality, consistent with precedent \cite{ttest, argaze}. Normality was assessed using the Shapiro-Wilk test. The Bonferroni correction was applied to adjust p-values, with a significance threshold of $\alpha = 0.05$. Outliers were identified using Tukey’s Fences, and Mauchly’s test of sphericity was conducted; when sphericity was violated, the Greenhouse-Geisser correction was applied.}

\section{Results}

\subsection{Task Completion Time}
\begin{revblock}
\begin{figure}[ht]
    \centering
    \includegraphics[width=1.0\linewidth]{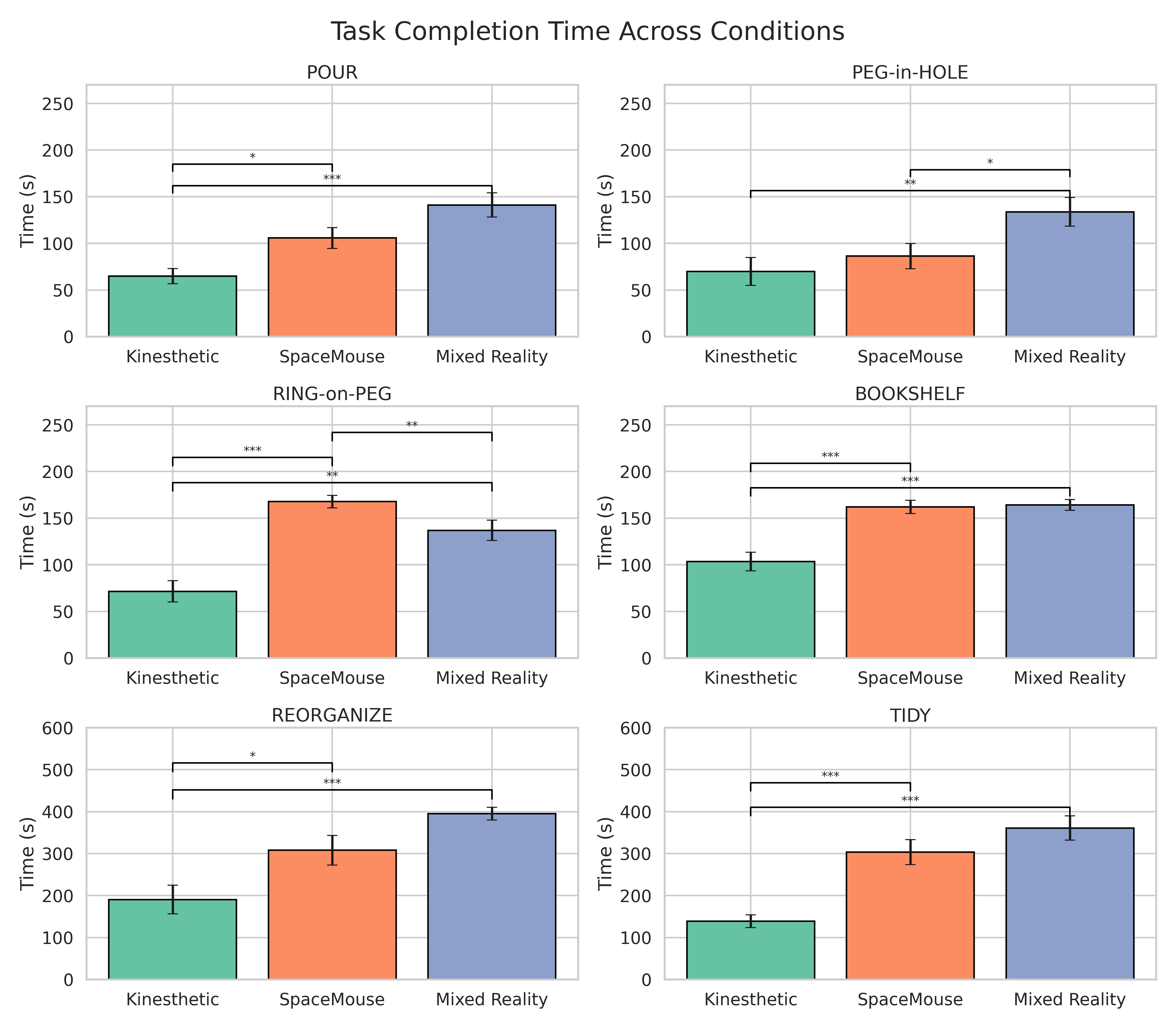}
    \caption{Bar plots illustrating task completion times across the six tasks. The sample size was N = 15 for the first four tasks and N = 10 for the final two tasks. Statistical significance is denoted by asterisks above the brackets: * for p < 0.05, ** for p < 0.01, and *** for p < 0.001.}
    \label{fig:time}
\end{figure}

\begin{table*}[ht]
\centering
\caption{Task completion time results. Repeated-measures ANOVA results and Bonferroni-corrected post-hoc comparisons for all tasks are shown. $F$ denotes the F-statistic with associated degrees of freedom (\textbf{df}). $p$ indicates the uncorrected significance value. $\eta_G^2$ represents the generalized effect size, and $p_{corr}$ reports the corrected p-value. Significant $p$-values ($p < .05$) are shown in \textbf{bold}.}
\label{tab:time_anova_posthoc}
\begin{tabular}{llcccc}
\toprule
\textbf{Task} & \textbf{Comparison} & \textbf{$F$ (df)} & \textbf{$p$} & \textbf{$\eta_G^2$} & \textbf{$p_{\text{corr}}$} \\
\midrule
\multirow{3}{*}{POUR} 
& Kinesthetic -- Mixed Reality & \multirow{3}{*}{$F(2,28)=13.23$} & \multirow{3}{*}{$<.001$} & \multirow{3}{*}{0.37} & \textbf{0.0004} \\
& Kinesthetic -- SpaceMouse    &  &  &  & \textbf{0.0383} \\
& Mixed Reality -- SpaceMouse  &  &  &  & 0.1132 \\
\midrule
\multirow{3}{*}{PEG-in-HOLE} 
& Kinesthetic -- Mixed Reality & \multirow{3}{*}{$F(2,28)=7.52$} & \multirow{3}{*}{0.002} & \multirow{3}{*}{0.20} & \textbf{0.0020} \\
& Kinesthetic -- SpaceMouse    &  &  &  & 1.0000 \\
& Mixed Reality -- SpaceMouse  &  &  &  & \textbf{0.0288} \\
\midrule
\multirow{3}{*}{RING-on-PEG} 
& Kinesthetic -- Mixed Reality & \multirow{3}{*}{$F(1.24,17.38)=27.45$} & \multirow{3}{*}{$<.001$} & \multirow{3}{*}{0.54} & \textbf{0.0066} \\
& Kinesthetic -- SpaceMouse    &  &  &  & \textbf{0.00001} \\
& Mixed Reality -- SpaceMouse  &  &  &  & \textbf{0.0081} \\
\midrule
\multirow{3}{*}{BOOKSHELF} 
& Kinesthetic -- Mixed Reality & \multirow{3}{*}{$F(2,26)=39.54$} & \multirow{3}{*}{$<.001$} & \multirow{3}{*}{0.50} & \textbf{0.0001} \\
& Kinesthetic -- SpaceMouse    &  &  &  & \textbf{0.00001} \\
& Mixed Reality -- SpaceMouse  &  &  &  & 1.0000 \\
\midrule
\multirow{3}{*}{REORGANIZE} 
& Kinesthetic -- Mixed Reality & \multirow{3}{*}{$F(2,18)=17.25$} & \multirow{3}{*}{$<.001$} & \multirow{3}{*}{0.47} & \textbf{0.0006} \\
& Kinesthetic -- SpaceMouse    &  &  &  & \textbf{0.0305} \\
& Mixed Reality -- SpaceMouse  &  &  &  & 0.1014 \\
\midrule
\multirow{3}{*}{TIDY} 
& Kinesthetic -- Mixed Reality & \multirow{3}{*}{$F(2,18)=19.91$} & \multirow{3}{*}{$<.001$} & \multirow{3}{*}{0.60} & \textbf{0.0006} \\
& Kinesthetic -- SpaceMouse    &  &  &  & \textbf{0.0009} \\
& Mixed Reality -- SpaceMouse  &  &  &  & 0.6334 \\
\bottomrule
\end{tabular}
\end{table*}

Across tasks, we observed statistically significant differences in task completion time, summarized in Table~\ref{tab:time_anova_posthoc}. Figure~\ref{fig:time} illustrates the average task completion times across conditions for each task. Overall trends indicate that participants generally completed tasks more quickly using the SpaceMouse compared to the Mixed Reality Interface, although statistically significant differences were observed only for the PEG-in-HOLE and RING-on-PEG tasks. Notably, these two tasks exhibited contrasting results: in the PEG-in-HOLE task, participants performed faster with the SpaceMouse, whereas in the RING-on-PEG task, participants performed faster with the Mixed Reality Interface.

The primary distinction between these tasks lies in their interaction requirements. The PEG-in-HOLE task demands a strictly top-down grasp and highly precise translational movements, while the RING-on-PEG task involves less stringent positional accuracy but requires substantial rotational motion. We hypothesize that the SpaceMouse becomes increasingly challenging to use as interactions move away from the initial reference frame, leading to less intuitive control mappings and reduced performance in tasks requiring large rotations. In contrast, the mixed reality interface better accommodates large rotational movements due to its visually grounded control paradigm; however, achieving precise, fine-grained motions requires greater user skill and control.
\end{revblock}

\subsection{Success}
\begin{revblock}
\begin{figure}[ht]
    \centering
    \includegraphics[width=1.0\linewidth]{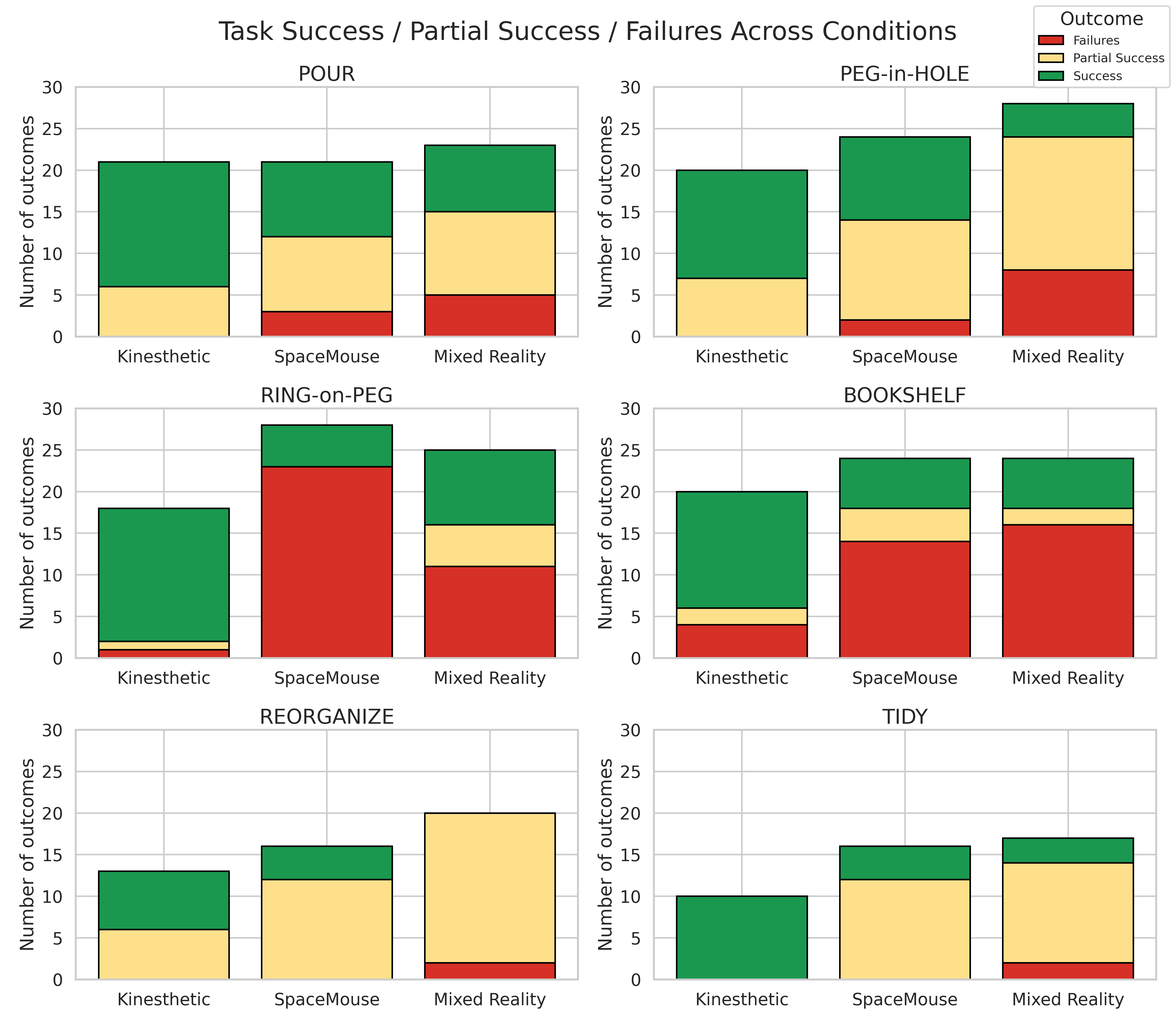}
    \caption{Counts of task outcomes aggregated across all participants for each task, including successful completions (green), partial successes (yellow), and failures (red).}
    \label{fig:success}
\end{figure}

Participants using the SpaceMouse experienced substantial difficulty in tasks requiring extensive rotation or navigation through constrained workspaces (RING-on-PEG and BOOKSHELF). While many participants were able to partially complete most tasks, the proportion of partial to complete successes increased markedly as task complexity and the number of required steps grew. Similar patterns were observed for the Mixed Reality interface; however, this method resulted in a greater number of failed trials for most tasks, leading to more overall attempts and generally fewer fully successful completions. As predicted, participants achieved the highest success rates using the Kinesthetic approach. This result demonstrates its role as a best-case performance benchmark for most users. Figure~\ref{fig:success} summarizes the success counts across conditions and Table ~\ref{tab:time_anova_posthoc} summarizes the statistical analysis.
\end{revblock}

\subsection{System Usability}
\begin{revblock}
\begin{figure}
    \centering
    \includegraphics[width=0.95\linewidth]{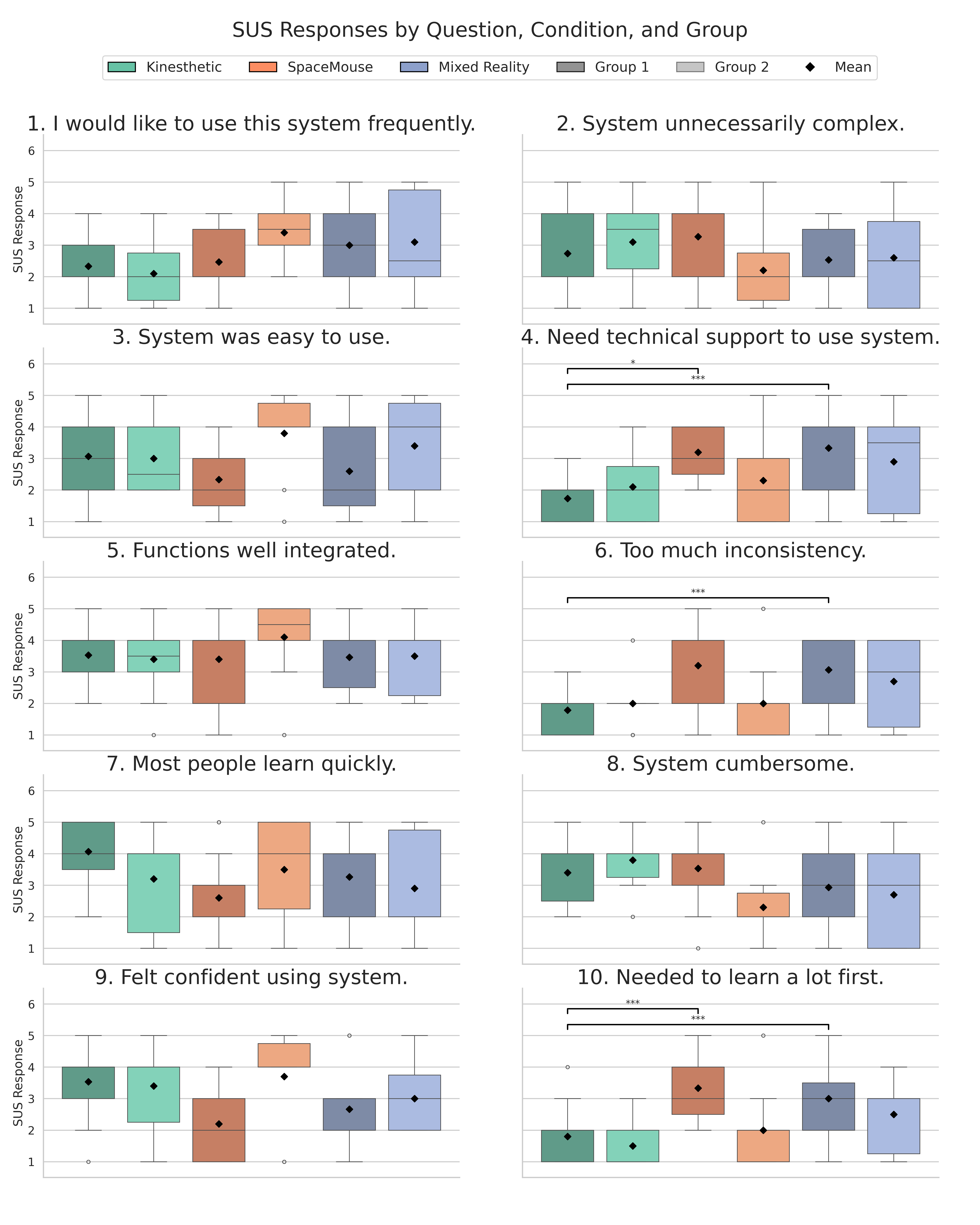}
    \caption{Box plots of \gls{sus} scores. Boxes show the 25th–75th percentiles, with the median indicated by a gray line and the mean by a diamond. Whiskers extend to 1.5× the interquartile range, and outliers are shown as unfilled circles. Brackets indicate statistically significant differences, with asterisks marking significance levels: p < 0.05 (*), p < 0.01 (**), and p < 0.001 (***).}
    \label{fig:sus}
\end{figure}

\begin{table*}
\small
\centering
\caption{SUS results per question from mixed ANOVA across Groups 1 and 2. The within-subject factor is Condition (Kinesthetic, Mixed Reality, SpaceMouse), the between-subject factor is Group (G1, G2), and significant post-hoc comparisons for Condition (Bonferroni-corrected) are indicated in \textbf{bold}. $F$ denotes the F-statistic with associated degrees of freedom (\textbf{df}). $p$ indicates the uncorrected significance value, and $\eta_p^2$ represents the partial effect size.}
\label{tab:sus_mixedanova}

\begin{tabularx}{\linewidth}{>{\raggedright\arraybackslash}X c c c c >{\raggedright\arraybackslash}X}
\toprule
\textbf{Question} & \textbf{Effect} & \textbf{$F$(df)} & \textbf{$p$} & \textbf{$\eta_p^2$} & \textbf{Post-hoc comparisons} \\
\midrule
\multirow{3}{*}{1. Would like to use frequently}
& Group        & $F(1,23)=0.77$ & 0.389 & 0.032 & -- \\
& Condition    & $F(2,48)=3.38$ & 0.043 & 0.128 & -- \\
& Interaction  & $F(2,46)=1.69$ & 0.196 & 0.068 & -- \\
\midrule
\multirow{3}{*}{2. System unnecessarily complex}
& Group        & $F(1,23)=0.33$ & 0.572 & 0.014 & -- \\
& Condition    & $F(2,48)=0.53$ & 0.594 & 0.022 & -- \\
& Interaction  & $F(2,46)=2.38$ & 0.104 & 0.094 & -- \\
\midrule
\multirow{3}{*}{3. System easy to use}
& Group        & $F(1,23)=4.72$ & 0.040 & 0.170 & -- \\
& Condition    & $F(2,48)=0.08$ & 0.921 & 0.004 & -- \\
& Interaction  & $F(2,46)=2.45$ & 0.098 & 0.096 & -- \\
\midrule
\multirow{3}{*}{4. Need technical support}
& Group        & $F(1,23)=0.98$ & 0.332 & 0.041 & -- \\
& Condition    & $F(2,48)=10.44$ & $<.001$ & 0.312 & K vs \gls{mr}: \textbf{<0.001}, K vs SM: \textbf{0.018} \\
& Interaction  & $F(2,46)=2.32$ & 0.110 & 0.092 & -- \\
\midrule
\multirow{3}{*}{5. Functions well integrated}
& Group        & $F(1,23)=0.35$ & 0.559 & 0.015 & -- \\
& Condition    & $F(2,48)=0.34$ & 0.717 & 0.014 & -- \\
& Interaction  & $F(2,46)=1.18$ & 0.318 & 0.049 & -- \\
\midrule
\multirow{3}{*}{6. Too much inconsistency}
& Group        & $F(1,23)=1.32$ & 0.263 & 0.057 & -- \\
& Condition    & $F(2,48)=7.51$ & 0.002 & 0.254 & K vs \gls{mr}: \textbf{0.001} \\
& Interaction  & $F(2,46)=3.08$ & 0.056 & 0.123 & -- \\
\midrule
\multirow{3}{*}{7. Most people learn quickly}
& Group        & $F(1,23)=0.11$ & 0.748 & 0.005 & -- \\
& Condition    & $F(2,48)=2.31$ & 0.111 & 0.091 & -- \\
& Interaction  & $F(2,46)=2.86$ & 0.067 & 0.111 & -- \\
\midrule
\multirow{3}{*}{8. System cumbersome}
& Group        & $F(1,23)=1.41$ & 0.247 & 0.058 & -- \\
& Condition    & $F(2,48)=2.61$ & 0.085 & 0.102 & -- \\
& Interaction  & $F(2,46)=3.07$ & 0.056 & 0.118 & -- \\
\midrule
\multirow{3}{*}{9. Felt confident using system}
& Group        & $F(1,23)=2.80$ & 0.108 & 0.108 & -- \\
& Condition    & $F(2,48)=3.51$ & 0.038 & 0.132 & -- \\
& Interaction  & $F(2,46)=3.87$ & 0.028 & 0.144 & -- \\
\midrule
\multirow{3}{*}{10. Needed to learn a lot first}
& Group        & $F(1,23)=4.70$ & 0.041 & 0.170 & -- \\
& Condition    & $F(2,48)=13.30$ & $<.001$ & 0.366 & K vs \gls{mr}, K vs SM: \textbf{<0.001} \\
& Interaction  & $F(2,46)=2.29$ & 0.112 & 0.091 & -- \\
\bottomrule
\end{tabularx}
\end{table*}

We report the \gls{sus} results across the entire participant population in Figure~\ref{fig:sus} on a per-question basis. Interestingly, no statistically significant differences were observed among the three methods for participants in Group 2, although trends suggest that these participants perceived the SpaceMouse as the most usable overall in the two long-horizon tasks. For Group 1, no statistically significant differences emerged between the two teleoperation methods. Table~\ref{tab:sus_mixedanova} shows the results from the mixed ANOVA, including group, condition, and interaction effects, for each SUS question.
\end{revblock}

\subsection{Task Load}
\begin{revblock}
\begin{figure}
    \centering
    \includegraphics[width=1.0\linewidth]{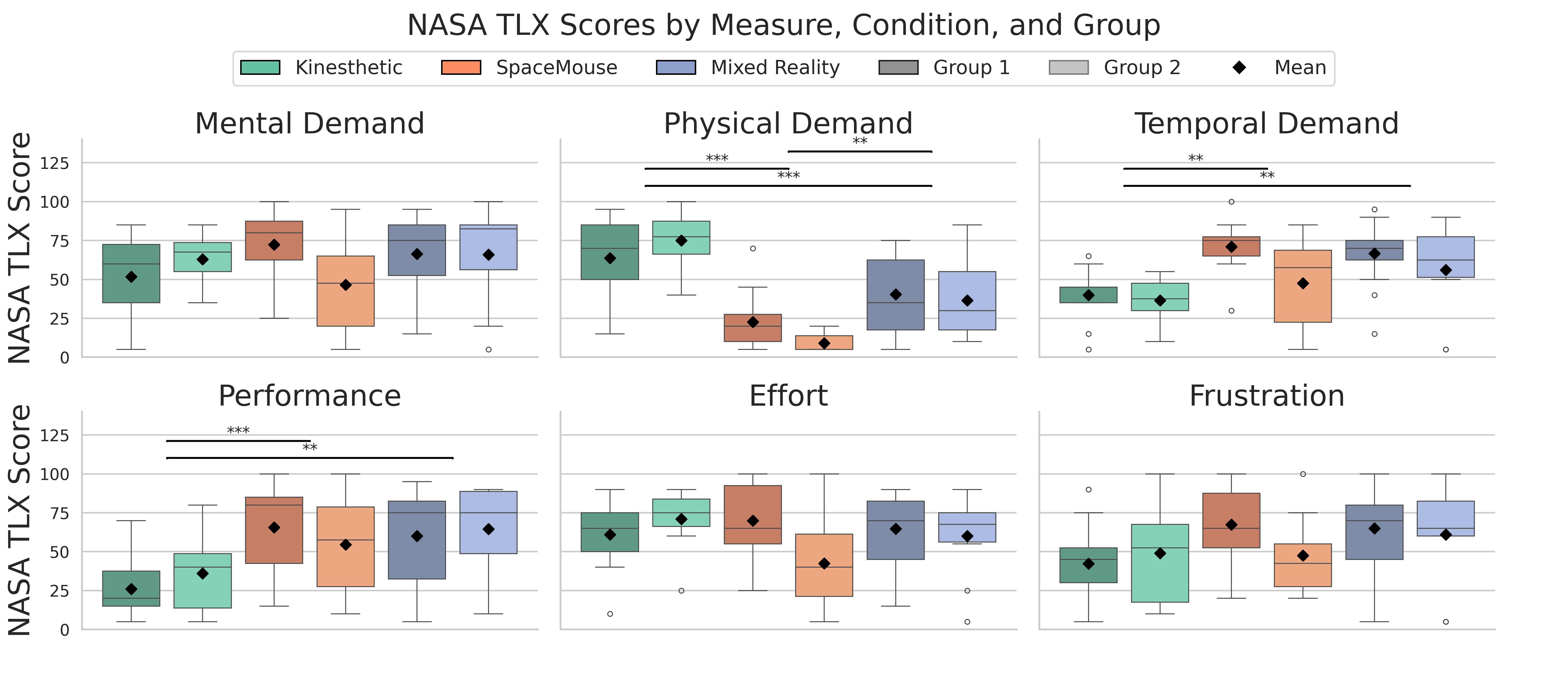}
    \caption{Box plots of NASA \gls{tlx} results. Lower scores are favorable across all measures. Asterisks denote statistical significance from post-hoc comparisons for Condition.}
    \label{fig:nasa}
\end{figure}

\begin{table*}
\small
\centering
\caption{NASA-TLX results from mixed ANOVA across Groups 1 and 2. The within-subject factor is Condition (Kinesthetic, Mixed Reality, SpaceMouse), the between-subject factor is Group (G1, G2), and significant post-hoc comparisons for Condition (Bonferroni-corrected) are indicated in \textbf{bold}. $F$ denotes the F-statistic with associated degrees of freedom (\textbf{df}). $p$ indicates the uncorrected significance value, and $\eta_p^2$ represents the partial effect size.}
\label{tab:nasa_mixedanova}

\begin{tabularx}{\linewidth}{>{\raggedright\arraybackslash}X c c c c >{\raggedright\arraybackslash}X}
\toprule
\textbf{Subscale} & \textbf{Effect} & \textbf{$F$(df)} & \textbf{$p$} & \textbf{$\eta_p^2$} & \textbf{Post-hoc comparisons} \\
\midrule
\multirow{3}{*}{Mental Demand}
& Group        & $F(1,23)=0.73$ & 0.403 & 0.03 & -- \\
& Condition    & $F(2,48)=1.03$ & 0.366 & 0.04 & -- \\
& Interaction  & $F(2,46)=3.54$ & 0.037 & 0.13 & -- \\
\midrule
\multirow{3}{*}{Physical Demand}
& Group        & $F(1,23)=0.17$ & 0.680 & 0.01 & -- \\
& Condition    & $F(2,48)=36.74$ & $<.001$ & 0.61 & K vs \gls{mr}, \gls{mr} vs SM: \textbf{0.001}, K vs SM: \textbf{<0.001} \\
& Interaction  & $F(2,46)=2.13$ & 0.130 & 0.08 & -- \\
\midrule
\multirow{3}{*}{Temporal Demand}
& Group        & $F(1,23)=4.29$ & 0.050 & 0.16 & -- \\
& Condition    & $F(2,48)=12.10$ & $<.001$ & 0.35 & K vs \gls{mr}, K vs SM: \textbf{0.001} \\
& Interaction  & $F(2,46)=1.69$ & 0.197 & 0.07 & -- \\
\midrule
\multirow{3}{*}{Performance}
& Group        & $F(1,23)=0.03$ & 0.872 & 0.00 & -- \\
& Condition    & $F(2,48)=12.26$ & $<.001$ & 0.35 & K vs \gls{mr}: \textbf{0.002}, K vs SM: \textbf{<0.001} \\
& Interaction  & $F(2,46)=1.07$ & 0.350 & 0.04 & -- \\
\midrule
\multirow{3}{*}{Effort}
& Group        & $F(1,23)=1.54$ & 0.227 & 0.06 & -- \\
& Condition    & $F(2,48)=0.44$ & 0.645 & 0.02 & -- \\
& Interaction  & $F(2,46)=4.12$ & 0.023 & 0.15 & -- \\
\midrule
\multirow{3}{*}{Frustration}
& Group        & $F(1,23)=0.81$ & 0.379 & 0.03 & -- \\
& Condition    & $F(2,48)=3.13$ & 0.053 & 0.12 & -- \\
& Interaction  & $F(2,46)=1.43$ & 0.250 & 0.06 & -- \\
\bottomrule
\end{tabularx}
\end{table*}

Finally, the NASA \gls{tlx} results revealed several statistically significant differences, specifically in the Physical Demand, Temporal Demand, Performance, and Frustration subscales. The average scores for each subscale are visualized in Figure~\ref{fig:nasa} and the statistical results for each subscale are summarized in Table \ref{tab:nasa_mixedanova}. The only measure showing significant differences across all three conditions was Physical Demand, with the SpaceMouse eliciting the lowest physical demand and the Kinesthetic method eliciting the highest physical demand across participants. For the remaining subscales, participants reported comparable workload levels between the two teleoperation methods.
\end{revblock}

\section{Discussion}

\begin{revblock}
Overall, the results highlight the limited effectiveness of non-experts in controlling complex robotic systems via teleoperation and reveal multiple areas for improvement. Many of these limitations emerged only under specific task demands, underscoring the need for comprehensive evaluations that encompass a diverse set of tasks. Notably, these findings quantify the performance gap between teleoperation methods and direct robot control: participants were approximately 1.5–2 times slower when completing tasks via teleoperation and up to four times less successful, often requiring additional trials. In demonstration contexts, this increased effort can contribute to user fatigue. We expect that other teleoperation approaches would exhibit similar trends, making it critical to reduce this performance gap—particularly in scenarios where kinesthetic teaching is impractical due to physical constraints or when users must maintain a safe distance from the workspace.

\subsection{Performance based on Prior Experience}
\begin{figure}
    \centering
    \includegraphics[width=1.0\linewidth]{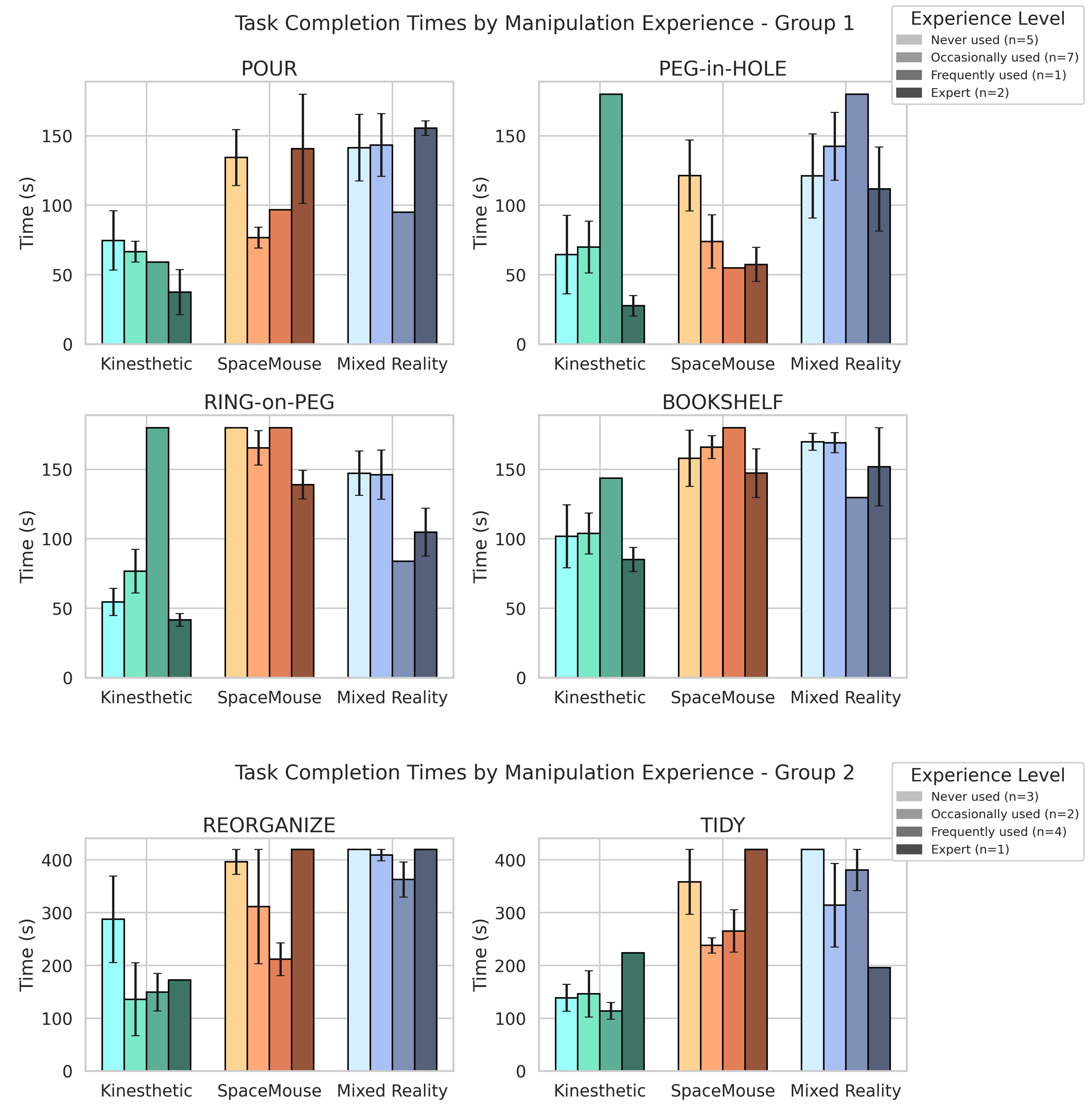}
    \caption{Task completion times based on prior experience in robotic manipulation.}
    \label{fig:time_manipulation_experience}
\end{figure}

As noted previously, the majority of participants in this study were non-experts; however, a subset of participants possessed prior expertise in specific experience categories, enabling limited comparisons between expert and non-expert performance. When analyzing task completion time results based on prior manipulation experience, we observed several instances in which non-experts and experts performed comparably. Figure \ref{fig:time_manipulation_experience} shows the average times based on this prior experience. For certain task–method combinations—for example, the Kinesthetic approach in the POUR task—performance trends aligned with expectations, with participants exhibiting greater experience demonstrating improved performance.

It is important to emphasize that extensive experience in robotic manipulation does not necessarily translate to familiarity with teleoperation interfaces. Given the small number of participants classified as strict experts, we refrain from drawing strong conclusions from these comparisons. Performance with the SpaceMouse generally reflected prior experience with the device, with more experienced users completing tasks more quickly across the majority of tasks. Additionally, participants with greater video gaming experience tended to perform better than their counterparts for both the SpaceMouse and Mixed Reality interfaces, likely due to enhanced spatial awareness and proficiency with complex control schemes. For further visualizations based on prior experience in other categories, we refer readers to the project website \cite{smith2025human_centered_teleop}.

\begin{table}[!]
\centering
\caption{Advantages of each teleoperation method, with themes, approximate number of supporting responses, and representative quotes.}
\label{tab:thematic_analysis}
\begin{tabular}{p{1.8cm} p{3cm} p{2cm} p{7cm}}
\toprule
\textbf{Method} & \textbf{Advantages} & \textbf{Responses (\#)} & \textbf{Representative Quotes} \\
\midrule
Kinesthetic & Direct and precise control & 15 & "You can control exactly what you want the kinova to do. You can manually move the joints anywhere and you have more control." \\
            & Tactile feedback and embodiment & 12 & "Familiar with real-life manipulation. Tactile/tangible change. Easy to recover when 'stuck'." \\
            & Learning and intuitiveness & 10 & "Easy and fast to learn. Directly see and feel the gripper's position. Very intuitive." \\
\midrule
SpaceMouse & Ease of use / Comfort & 14 & "The SpaceMouse was the most comfortable." \\
           & Fine-grained control & 12 & "Very fine-grained control. Smooth motions, low effort to move and orient the robot." \\
           & Efficiency / Low cognitive load & 10 & "Easy to use. Not too mentally intense. Consistent among different tasks." \\
\midrule
Mixed Reality & Reduced physical effort & 15 & "No physical effort. Somehow had to think less about self-collisions/joint constraints when I was just focusing on getting the end effector pose right." \\
                   & Enhanced spatial awareness / planning & 13 & "I liked that I was able to see an image of the goal state before I actually reached the goal state." \\
                   & Intuitive goal-oriented operation & 11 & "The AR approach made it easy to set an end pose and let the robot figure out how to get there. There is less planning with which joints to move to avoid singularities." \\
\bottomrule
\end{tabular}
\end{table}

In the post-experiment questionnaire, participants were asked to describe the advantages of each method relative to the other two. Table~\ref{tab:thematic_analysis} presents a thematic analysis of these responses. The following discussion interprets the results in the context of users’ ability to do, learn, and understand within the teleoperation tasks.

\subsection{Helping Users Do} Participants identified two primary attributes of the Kinesthetic approach that enabled superior objective performance. First, direct joint-level control allowed users to manipulate individual joints explicitly, rather than relying on end-effector control and inverse kinematics solvers, which can produce configurations that are difficult to recover from or maneuver out of. Across both teleoperation methods, participants frequently struggled near kinematic singularities and joint limits, resulting in prolonged periods of feeling "stuck" before discovering how to recover from these configurations. Second, haptic feedback provided users with an intuitive sense of feasible motion directions based on the robot’s underlying kinematics. The SpaceMouse was perceived as the interface that best supported users’ ability to act, which matched our hypothesis. Its compact form factor and ease of use made participants feel more comfortable when moving the robot, regardless of whether the resulting motion actually matched their intent. Furthermore, this device allowed for very fine-grained control, allowing participants to make small adjustments to the end-effector pose very easily. This general ease of motion had a pronounced impact on participants’ perceived workload. Participants also highlighted the reduced physical effort associated with the Mixed Reality interface, further underscoring the importance of enabling users to maintain a direct and intuitive sense of control over the robot’s motion. Moreover, participants described voice commands for gripper control as convenient, intuitive, and easy to use, underscoring the promise of voice-based interaction as an effective control modality for robots in household environments. Although performance and user preference may be influenced by ambient noise, the strong positive reception suggests that voice commands offer a natural and accessible means of interaction for everyday robotic tasks.

\subsection{Helping Users Learn} Several participants reported that the Kinesthetic approach was easy to learn and highly intuitive. This perception stemmed from their ability to directly observe the entire manipulator within the workspace while simultaneously receiving joint-level haptic feedback. In contrast, the two teleoperation approaches presented greater learning challenges, albeit for different reasons.

For the Mixed Reality method, we observed that the underlying \gls{mr} interaction paradigm was difficult for many participants to grasp, despite prior experience with \gls{vr}/\gls{ar}. Notably, none of the participants had previously used the HoloLens 2. Participants primarily struggled with the pinch-and-grab interaction used to manipulate virtual objects, which, while not specific to our interface design, is a fundamental interaction mechanism of the \gls{mr} headset itself. Although increased familiarity with the HoloLens 2 may lead to improved objective performance with \gls{mr}-based teleoperation interfaces, the requirement for non-expert users to first learn a specialized \gls{mr} interaction paradigm is an important consideration, as it may restrict the potential user pool and increase barriers to accessibility.

The SpaceMouse also posed learning challenges, largely due to changing control frames induced by rotations. As users moved beyond the initial reference frame, the mapping between device input and end-effector motion continuously changed. Consequently, users who became accustomed to a particular mapping were required to relearn the control relationships after rotations, often resorting to trial-and-error end-effector movements unrelated to the task objectives. The RING-on-PEG results highlight the impact of rotation-related unintuitiveness on user performance.

\subsection{Helping Users Understand} Many participants identified distinctive features of the Mixed Reality interface as beneficial for understanding robot behavior, despite initial difficulties adapting to the interaction paradigm. In particular, the ability to visualize the next desired pose or preview the robot’s intended motion was viewed as a significant advantage of the Mixed Reality method—one that is not inherently available in the other control approaches. Participants reported that end-effector visualization enabled them to specify goal poses more easily, while allowing the inverse kinematics solver to determine the corresponding joint configurations.

Additionally, the virtual gripper enhanced participants’ spatial awareness by providing a clearer (view-independent) understanding of how the robot would move through the physical workspace before the arm executed the motion. This anticipatory visualization helped users reason about feasibility and collision avoidance from various angles and positions in space.

As noted previously, participants using the Kinesthetic approach relied instead on tactile feedback to develop a similar understanding of the robot’s motion capabilities, highlighting complementary strengths between physical and visually augmented interaction modalities.

\subsection{Moving Forward}
The results presented in this study provide insights into the advantages and limitations of two widely used robot control schemes, both in terms of objective performance and subjective user perceptions. They also demonstrate that certain design elements, while beneficial in user-centered teleoperation, may still be insufficient to achieve the performance levels attainable with alternative methods. Given the observed performance gap between teleoperation and direct manipulation, it is critical to continue evaluating existing interfaces using comparable methodologies—assessing them across diverse systems and complex tasks—to ensure that users are able to do, learn, and understand simultaneously, rather than excelling in only one of these dimensions.

Although users encountered challenges with the \gls{mr} interface, we posit that \gls{mr}-based interfaces—beyond this specific implementation—have significant potential to facilitate doing, learning, and understanding, particularly when integrated with \gls{ai}. Developing interfaces that can be customized to individual user preferences may provide substantial advantages in teleoperation contexts. A key benefit of \gls{mr} lies in its ability to offer configurable interface options, enabling users to tailor control methods, gestures, and visualizations to their needs. For instance, the interface used in this study provided three different methods for controlling the gripper, and all holograms could be manipulated via near or far interactions. By allowing users to select the interaction modes that best suit their requirements, such customization accommodates diverse user preferences while preserving interface simplicity. While these options were predefined by the designer, \gls{ai} integration could further expand the range of adaptive, user-specific configurations.
\end{revblock}

\begin{figure*}[h]
\centering
\begin{tabular}{lll}
\includegraphics[scale=0.45]{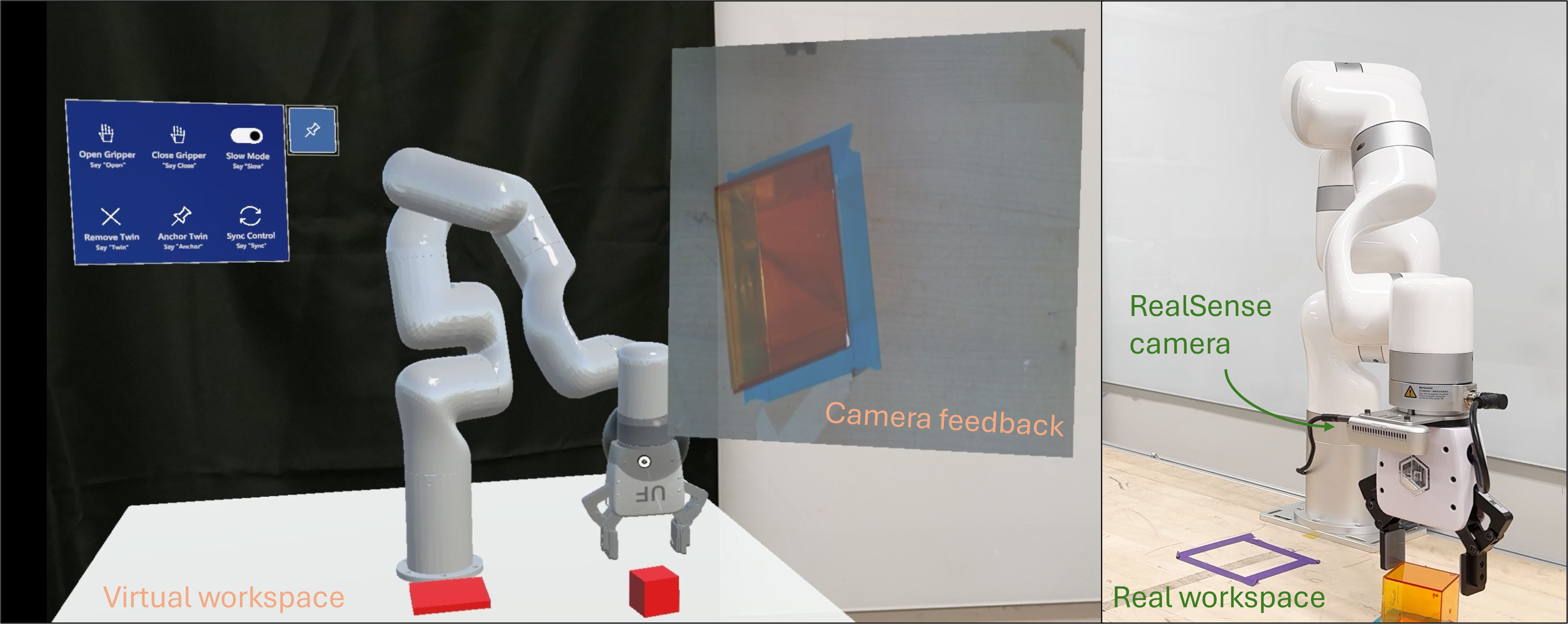}
\end{tabular}
\caption{A robot and environment modeled in \gls{mr} (left) and the corresponding real-world environment (right). }
\label{remote_teleop}
\end{figure*}

\rev{Moreover, in scenarios where collocated teleoperation is not feasible, \gls{mr} interfaces support 3D visualizations that enhance spatial understanding.} Figure \ref{remote_teleop} illustrates one such scenario, in which the operator interacts with a virtual replica of the physical environment, supplemented by the robot’s wrist camera feed. Compared to conventional 2D first- or third-person camera views, 3D visualizations make navigation and control of the robot more intuitive. The realism and immersion of the virtual environment could be further improved using advanced 3D reconstruction techniques, such as Gaussian Splatting or Neural Radiance Fields (NeRFs), although implementing these approaches would require additional computational resources.

\subsection{Limitations} 
\begin{revblock}
As stated previously, this study is not exhaustive. Given the wide range of robot control systems that have been proposed, conducting similar analyses across many approaches is important. However, including additional methods in a single study would substantially increase participant fatigue. We therefore selected three representative control methods that capture common interaction paradigms, and we leave comparisons with other approaches for future work.

Additionally, this study focused on stationary single-arm manipulators. As robotic systems increase in complexity and degrees of freedom, real-time control becomes even more challenging. Ensuring that non-expert users can effectively control bimanual, humanoid, or mobile robots is equally critical, particularly as such systems are likely to operate in personal spaces. We encourage future studies to extend this type of comparative analysis to more complex robotic platforms.

Finally, interface implementation details play a critical role in user performance, particularly for mixed reality interfaces. We proactively addressed these factors through a preliminary pilot study that isolated and resolved tuning and fidelity issues prior to the main evaluation. Nevertheless, more rigorous and controlled comparisons of alternative \gls{mr} interface designs are warranted, and we leave such investigations to future work.
\end{revblock}


\section{Conclusion}

\rev{In this study, we analyzed two teleoperation methods—SpaceMouse Teleoperation and a \gls{mr} teleoperation interface—comparing their performance to a widely used non-teleoperation approach for robotic control: Kinesthetic Teaching. All three methods were evaluated through a comprehensive user study involving two robotic platforms and six challenging manipulation tasks. The SpaceMouse and the \gls{mr} interface performed comparably across the set of tasks, with statistically significant differences in task completion time observed for only two tasks, and overall success rates declining as task complexity increased. Qualitative findings mirrored these trends, with the most pronounced differences reflected in the Physical Demand imposed by each method. Further analysis identified the specific attributes of each interface that influenced participants’ ability to perform, learn, and understand in the context of these tasks.}

\rev{Within the broader context of designing teleoperation interfaces that approximate the performance of the direct manipulation, this study quantified the performance limitations of teleoperation methods such as the SpaceMouse and \gls{mr} interfaces. The findings indicate that physical devices, haptic feedback, and/or direct joint-level control may be critical for achieving high-performance teleoperation, complementing visual feedback and interface personalization. We encourage researchers—particularly those developing autonomous robots for deployment in personal spaces—to design and rigorously evaluate teleoperation systems for non-expert users, as such tools will become increasingly important as robots are integrated into everyday environments.}

\printbibliography










\end{document}